%% file: ShortSurvey.tex
\documentclass[twoside,11pt]{article}

\usepackage{jair, theapa, rawfonts}
\usepackage{filecontents}
\usepackage[round]{natbib}
\usepackage{amsmath}
\usepackage[utf8x]{inputenc}
\usepackage{bibentry}
\usepackage{courier}
\usepackage{xspace}
\usepackage[encapsulated]{CJK}
\usepackage{url}
\usepackage{float}
\usepackage{graphicx}

\usepackage{amsfonts}

\newcommand{\zh}[1]{\begin{CJK}{UTF8}{bsmi}#1\end{CJK}}
\newcommand{\jp}[1]{\begin{CJK}{UTF8}{min}#1\end{CJK}}

\def\bibname{\noindent{\textbf{References}}}

\input{macros}

\jairheading{}{}{}{10/2018}{- Under Review}
\ShortHeadings{Automatic Language Identification in Texts: A Survey}
{Jauhiainen, Lui, Zampieri, Baldwin, \& Lind{\'e}n}
\firstpageno{1}

\title{Automatic Language Identification in Texts: A Survey}
\author{\name Tommi Jauhiainen \email tommi.jauhiainen@helsinki.fi \\
  \addr Department of Digital Humanities\\
  The University of Helsinki\\
  \AND
  \name Marco Lui \email saffsd@gmail.com \\
  \addr School of Computing and Information Systems\\
  The University of Melbourne\\
  \AND
  \name Marcos Zampieri \email m.zampieri@wlv.ac.uk \\
  \addr Research Institute in Information and Language Processing\\
  University of Wolverhampton\\
  \AND
  \name Timothy Baldwin \email tb@ldwin.net\\
  \addr School of Computing and Information Systems\\
  The University of Melbourne\\
  \AND
  \name Krister Lind{\'e}n \email krister.linden@helsinki.fi\\
  \addr Department of Digital Humanities\\
  The University of Helsinki\\
 }

\begin{document}

\maketitle

\begin{abstract}
  Language identification (``\langid'') is the problem of determining the
  natural language that a document or part thereof is written
  in. Automatic \langid has been extensively researched for over fifty years.
  Today, \langid is a key part of many text processing pipelines, as text
  processing techniques generally assume that the language of the input
  text is known. Research in this area has recently been especially active.
  This article provides a brief history of \langid research, and an
  extensive survey of the features and methods used in the \langid literature.
  We describe the features and methods using a unified notation, to make
  the relationships between methods clearer.
  We discuss evaluation methods, applications of \langid, as well as
  \emph{off-the-shelf} \langid systems that do not require training by
  the end user. Finally, we identify open issues, survey the work to
  date on each issue, and propose future directions for research in
  \langid.
\end{abstract}

\section{Introduction}
\label{sec:intro}

Language identification (``\langid'') is the task of determining the natural
language that a document or part thereof is written in. Recognizing text
in a specific language comes naturally to a human reader familiar with
the language. \tabref{intro:langid} presents excerpts from Wikipedia articles 
in different languages on the topic of Natural Language Processing (``NLP''), labeled 
according to the language they are written in. Without referring to the labels, readers of this article will certainly have recognized at least one language in \tabref{intro:langid}, and many are likely to be able to identify all the languages therein.

\begin{table}[t]
  \centering
  \begin{tabular}{lp{12cm}}
\hline
  English & Natural language processing is a field of computer science, artificial intelligence, and linguistics concerned with the interactions between computers and human (natural) languages.\\[2mm]
  Italian & L'Elaborazione del linguaggio naturale è il processo di trattamento automatico mediante un calcolatore elettronico delle informazioni scritte o parlate nel linguaggio umano o naturale.\\[2mm]
  Chinese & \zh{ 自然語言處理是人工智慧和語言學領域的分支學科。} \\[2mm]
  Japanese & \jp{自然言語処理は、人間が日常的に使っている自然言語をコンピュータに処理させる一連の技術であり、人工知能と言語学の一分野である。}\\[2mm]
\hline
  \end{tabular}
\caption{Excerpts from Wikipedia articles on NLP in different languages.}
\label{tab:intro:langid}
\end{table}

Research into \langid aims to mimic this human ability to recognize
specific languages. Over the years, a number of computational approaches
have been developed that, through the use of specially-designed
algorithms and indexing structures, are able to infer the language being
used without the need for human intervention. The capability
of such systems could be described as super-human: an average person may be able to
identify a handful of languages, and a trained linguist or translator
may be familiar with many dozens, but most of us will have, at some point,
encountered written texts in languages they cannot place. However,
\langid research aims to develop systems that are able to identify
\textit{any} human language, a set which numbers in the thousands
\citep{simons1}.

In a broad sense, \langid applies to any modality of language, including
speech, sign language, and handwritten text, and is relevant for all means of
information storage that involve language, digital or otherwise.
However, in this survey we limit the scope of our discussion to \langid
of written text stored in a digitally-encoded form.

Research to date on \langid has traditionally focused on \textit{monolingual} 
documents \citep{hughes1} (we discuss \langid for multilingual documents in
\secref{openissues:multilingual}).  In monolingual \langid, the task is to assign 
each document a unique language label.  Some work has reported near-perfect accuracy 
for \langid of large documents in a small number of languages,
prompting some researchers to label it a ``solved task'' \citep{mcnamee1}.
However, in order to attain such accuracy, simplifying assumptions have to be 
made, such as the aforementioned monolinguality of each document, as well as assumptions about 
the type and quantity of data, and the number of languages considered.

The ability to accurately detect the language that a document is written in is
an enabling technology that increases accessibility of data and has
a wide variety of applications.  For example, presenting information in a
user's native language has been found to be a critical factor in attracting
website visitors \citep{Kralisch:Mandl:2006}. Text processing techniques
developed in natural language processing and Information Retrieval (``IR'')
generally presuppose that the language of the input text is known, and many
techniques assume that all documents are in the same language. In order to
apply text processing techniques to real-world data, automatic \langid is used
to ensure that only documents in relevant languages are subjected to further
processing.  In information storage and retrieval, it is common to index
documents in a multilingual collection by the language that they are written
in, and \langid is necessary for document collections where the languages of
documents are not known a-priori, such as for data crawled from the World Wide
Web. Another application of \langid that predates computational methods is the
detection of the language of a document for routing to a suitable translator.
This application has become even more prominent due to the advent of Machine
Translation (``MT'') methods: in order for MT to be applied to translate a document
to a target language, it is generally necessary to determine the source
language of the document, and this is the task of \langid. \langid also plays a
part in providing support
for the documentation and use of low-resource languages. One area where
\langid is frequently used in this regard is in linguistic corpus creation,
where \langid is used to process targeted web crawls to collect text resources
for low-resource languages. 

A large part of the motivation for this article is the observation that
\langid lacks a ``home discipline'', and as such, the literature is
fragmented across a number of fields, including NLP, IR, machine learning, data mining, social
medial analysis, computer science education, and systems science. This
has hampered the field, in that there have been many instances of
research being carried out with only partial knowledge of other work on
the topic, and the myriad of published systems and datasets.

Finally, it should be noted that this survey does not make a distinction
between languages, language varieties, and dialects. Whatever
demarcation is made between languages, varieties and dialects, a \langid system is
trained to identify the associated document classes. Of course, the more similar two classes are, the
more challenging it is for a \langid system to discriminate between
them. Training a system to discriminate between similar languages such
as Croatian and Serbian \citep{ljubesic4}, language varieties like
Brazilian and European Portuguese \citep{zampieri4}, or a set of Arabic
dialects \citep{zampieri8} is more challenging than training systems to
discriminate between, for example, Japanese and Finnish. Even so, as
evidenced in this article, from a computational perspective, the
algorithms and features used to discriminate between languages, language
varieties, and dialects are identical.

\section{\langid as Text Categorization}
\label{sec:intro:textcategorization}

\langid is in some ways a special case of text categorization, and previous
research has examined applying standard text categorization methods to
\langid \citep{cavnar1,elworthy1}.

\cite[Section 2.1]{sebastiani1} provides a definition of text 
categorization, which can be summarized as the task of mapping a document
onto a pre-determined set of classes. This is a very broad definition, 
and indeed one that is applicable to a wide variety of tasks, amongst which 
falls modern-day \langid. The archetypal text categorization task
is perhaps the classification of newswire articles according to the topics that 
they discuss, exemplified by the Reuters-21578 dataset 
\citep{Debole:Sebastiani:2005}. However, \langid has particular characteristics that 
make it different from typical text categorization tasks: 
\begin{enumerate}
\item Text categorization tends to use statistics about the frequency of words 
  to model documents, but for \langid purposes there is no 
  universal notion of a \textit{word}: \langid must cater for 
  languages where whitespace is not used to denote word
  boundaries. Furthermore, the determination of the appropriate word
  tokenization strategy for a given document presupposes knowledge of
  the language the document is written in, which is exactly what we
  assume we \textit{don't} have access to in \langid.
\item In text categorization tasks, the set of labels usually only applies to a 
  particular dataset. For example, it is not meaningful to ask which of the 
  Reuters-21578 labels is applicable to the abstract of a biomedical journal 
  article. However, in \langid there is a clear notion of 
  language that is independent of domain: it is possible to recognize that a 
  text is in English regardless of whether it is from a biomedical
  journal, a microblog post, or a newspaper article.
\item In \langid, classes can be somewhat multi-modal, in that text in the same 
  language can sometimes be written with different orthographies and stored in 
  different encodings, but correspond to the same class.
\item In \langid, labels are non-overlapping and mutually exclusive,
  meaning that a text can only be written in one language. This does not
  preclude the existence of multilingual documents which contain text in
  more than one language, but when this is the case, the document can
  always be uniquely divided into monolingual segments. This is in
  contrast to text categorization involving multi-labeled documents,
  where it is generally not possible to associate specific segments of
  the document with specific labels.
\end{enumerate}
These distinguishing characteristics present unique challenges and offer 
particular opportunities, so much so that research in \langid has generally 
proceeded independently of text categorization research. 
In this survey, we will examine the common themes and ideas that underpin 
research in \langid. We begin with a brief history of research 
that has led to modern \langid (\secref{history}), and then proceed 
to review the literature, first introducing the mathematical notation used in the article (\secref{notation}), and then providing synthesis and analysis of existing 
research, focusing specifically on the representation of text 
(\secref{features}) and the learning 
algorithms used (\secref{methods}).
We examine the methods for evaluating the quality of the systems (\secref{evaluation}) as well as the areas where 
\langid has been applied (\secref{applications}), and 
then provide an overview of ``off-the-shelf'' \langid systems 
(\secref{ots}). We conclude the survey with a discussion of the open 
issues in \langid (\secref{openissues}), enumerating 
issues and existing efforts to address them, as well as charting the 
main directions where further research in \langid is required.

\subsection{Previous Surveys}
\label{sec:into:previous}

Although there are some dedicated survey articles, these tend to be
relatively short; there have not been any comprehensive surveys of
research in automated LI of text to date. The largest survey so far can
be found in the literature review of Marco Lui's PhD thesis \citep{lui7}, which
served as an early draft and starting point for the current
article. \cite{zampieri11} provides a historical overview of language
identification focusing on the use of \ngram language
models. \cite{qafmolla1} gives a brief overview of some of the methods
used for \langid, and \cite{garg1} provide a review of some of the
techniques and applications used previously. \cite{shashirekha1} gives
a short overview of some of the challenges, algorithms and available
tools for \langid. \cite{juola1} provides a brief summary of \langid,
how it relates to other research areas, and some outstanding challenges,
but only does so in general terms and does not go into any detail about
existing work in the area. Another brief article about \langid is
\cite{Muthusamy:Spitz:1997}, which covers \langid both of spoken
language as well as of written documents, and also discusses \langid of
documents stored as images rather than digitally-encoded text.

\section{A Brief History of \langid}
\label{sec:history}

\langid as a task predates computational methods -- the earliest
interest in the area was motivated by the needs of translators, and
simple manual methods were developed to quickly identify documents in
specific languages. The earliest known work to describe a functional
\langid program for text is by \cite{mustonen1}, a statistician, who
used multiple discriminant analysis to teach a computer how to
distinguish, at the word level, between English, Swedish and
Finnish. Mustonen compiled a list of linguistically-motivated
character-based features, and trained his language identifier on 300
words for each of the three target languages. The training procedure
created two discriminant functions, which were tested with 100 words for
each language. The experiment resulted in 76\% of the words being
correctly classified; even by current standards this percentage would be
seen as acceptable given the small amount of training material, although
the composition of training and test data is not clear, making the
experiment unreproducible.

In the early 1970s, \cite{nakamura1} considered the problem of
automatic \langid. According to \cite{rau1} and the available abstract
of Nakamura's article,\footnote{We were unable to obtain the original
  article, so our account of the paper is based on the abstract and
  reports in later published articles.} his language identifier was able
to distinguish between 25 languages written with the Latin alphabet. As
features, the method used the occurrence rates of characters and words
in each language. From the abstract it seems that, in addition to the
frequencies, he used some binary presence/absence features of particular
characters or words, based on manual \langid.

\cite{rau1} wrote his master's thesis ``Language Identification by
Statistical Analysis'' for the Naval Postgraduate School at Monterey,
California. The continued interest and the need to use \langid of text
in military intelligence settings is evidenced by the recent articles
of, for example, \cite{rehiman2}, \cite{rowe1}, \cite{tratz1}, and
\cite{voss2}. As features for \langid, \cite{rau1} used, e.g., the relative
frequencies of characters and character bigrams. With a
majority vote classifier ensemble of seven classifiers using
Kolmogor-Smirnov's Test of Goodness of Fit and Yule's characteristic ($K$),
he managed to achieve 89\% accuracy over 53 characters when
distinguishing between English and Spanish. His thesis actually includes
the identifier program code (for the IBM System/360 Model 67 mainframe),
and even the language models in printed form.

Much of the earliest work on automatic \langid was focused on
identification of spoken language, or did not make a distinction between
written and spoken language. For example, the work of \cite{house1} is
primarily focused on \langid of spoken utterances, but makes a broader
contribution in demonstrating the feasibility of \langid on the basis of
a statistical model of broad phonetic information. However, their
experiments do not use actual speech data, but rather ``synthetic'' data
in the form of phonetic transcriptions derived from written text.

Another subfield of speech technology, speech synthesis, has also
generated a considerable amount of research in the \langid of text,
starting from the 1980s. In speech synthesis, the need to know the
source language of individual words is crucial in determining how they
should be pronounced. \cite{church1} uses the relative frequencies of
character trigrams as probabilities and determines the language of words
using a Bayesian model. Church explains the method -- that has since
been widely used in LI -- as a small part of an article concentrating
on many aspects of letter stress assignment in speech synthesis, which
is probably why \cite{beesley1} is usually attributed to being the one
to have introduced the aforementioned method to \langid of text. As
Beesley's article concentrated solely on the problem of LI, this single
focus probably enabled his research to have greater visibility. The role
of the program implementing his method was to route documents to MT
systems, and Beesley's paper more clearly describes what has later come
to be known as a character \ngram model. The fact that the distribution
of characters is relatively consistent for a given language was already
well known.

The highest-cited early work on automatic \langid is
\cite{cavnar1}. Cavnar and Trenkle's method (which we describe in
detail in \secref{outofplace}) builds up per-document and per-language
profiles, and classifies a document according to which language profile it
is most similar to, using a rank-order similarity metric.  They evaluate
their system on 3478 documents in eight languages obtained from USENET
newsgroups, reporting a best overall \langid accuracy of 99.8\%. Gertjan
van Noord produced an implementation of the method of Cavnar and Trenkle
named \textcat, which has become eponymous with the method
itself. \textcat is packaged with pre-trained models for a
number of languages, and so it is likely that the strong results reported
by Cavnar and Trenkle, combined with the ready availability of an
``off-the-shelf'' implementation, has resulted in the exceptional
popularity of this particular method.  \cite{cavnar1} can be considered
a milestone in automatic \langid, as it popularized the use of automatic
methods on character \ngram models for \langid, and to date the method
is still considered a benchmark for automatic \langid.

\section{On Notation}
\label{sec:notation}

This section introduces the notation used throughout this article to
describe \langid methods. We have translated the notation in the
original papers to our notation, to make it easier to see the similarities
and differences between the \langid methods presented in the
literature. The formulas presented could be used to implement language
identifiers and re-evaluate the studies they were originally presented
in.

A corpus \(C\) consists of individual tokens \(u\) which may be bytes,
characters or words. \(C\) is comprised of a finite sequence of individual
tokens, \(u_{1}, ..., u_{l_C}\). The total count of individual tokens
\(u\) in \(C\) is denoted by \(l_C\). In a corpus \(C\) with
non-overlapping segments \(S\), each segment is referred to as
\(C_{s}\), which may be a short document or a word or some other way of
segmenting the corpus. The number of segments is denoted as \(l_S\).

A feature \(f\) is some countable characteristic of the corpus
\(C\). When referring to the set of all features \(F\) in a corpus
\(C\), we use \(C^{F}\), and the number of features is denoted by
\(l_{C^{F}}\). A set of unique features in a corpus \(C\) is denoted by
\(U(C)\). The number of unique features is referred to as
\(|U(C)|\). The count of a feature \(f\) in the corpus \(C\) is referred
to as \(c(C, f)\). If a corpus is divided into segments \(S\), the count
of a feature \(f\) in \(C\) is defined as the sum of counts over the
segments of the corpus, i.e. \(c(C,f) = \sum_{s=1}^{l_S}
c(C_{s},f)\). Note that the segmentation may affect the count of a
feature in \(C\) as features do not cross segment borders.

A frequently-used feature is an \ngram, which consists of a contiguous
sequence of \(n\) individual tokens. An \ngram starting at position
\(i\) in a corpus segment is denoted \(u_{i,...,i+n-1}\), where
positions \(i+1,...,i+n-1\) remain within the same segment of the corpus
as \(i\). If \(n=1\), \(f\) is an individual token. When referring to
all \ngrams of length \(n\) in a corpus \(C\), we use \(C^{n}\) and the
count of all such \ngrams is denoted by \(l_{C^{n}}\). The count of an
\ngram \(f\) in a corpus segment \(C_s\) is referred to as \(c(C_s, f)\)
and is defined by \eqnref{count}:

\begin{equation}
c(C_s,f)= \sum_{i=1}^{l_{C_s}+1-n} \left\{ 
  \begin{array}{l l}
  1 & \quad \text{, if } f = u_{i,...,i-1+n}\\
  0 & \quad \text{, otherwise}
  \end{array} \right.\\
\label{eqn:count}
\end{equation}
The set of languages is \(G\), and \(l_G\) denotes the number of
languages. A corpus \(C\) in language \(g\) is denoted by \(C_{g}\). A
language model \(O\) based on \(C_g\) is denoted by \(O(C_g)\). The
features given values by the model \(O(C_g)\) are the domain
\(dom(O(C_g))\) of the model. In a language model, a value \(v\) for the
feature \(f\) is denoted by \(v_{C_g}(f)\). For each potential language
\(g\) of a corpus \(C\) in an unknown language, a resulting score
\(R(g,C)\) is calculated. A corpus in an unknown language is also
referred to as a test document.

\subsection{An Archetypal Language Identifier}
\label{sec:textcat}

The design of a supervised language identifier can generally be deconstructed into four key steps:

\begin{enumerate}
  \item A representation of text is selected
  \item A model for each language is derived from a training corpus of
    labelled documents
  \item A function is defined that determines the similarity between a document and each language
  \item The language of a document is predicted based on the highest-scoring model
\end{enumerate}

\subsection{On the Equivalence of Methods}

The theoretical description of some of the methods leaves room for interpretation on how to implement them. \cite{cormen1} define an algorithm to be any well-defined computational procedure. \cite{yanofsky1} introduces a three-tiered classification where programs implement algorithms and algorithms implement functions. The examples of functions given by \cite{yanofsky1}, \emph{sort} and \emph{find max} differ from our \emph{identify language} as they are always solvable and produce the same results. In this survey, we have considered two methods to be the same if they always produce exactly the same results from exactly the same inputs. This would not be in line with the definition of an algorithm by \cite{yanofsky1}, as in his example there are two different algorithms \emph{mergesort} and \emph{quicksort} that implement the function \emph{sort}, always producing identical results with the same input. What we in this survey call a method, is actually a function in the tiers presented by \cite{yanofsky1}.

\section{Features}
\label{sec:features}

In this section, we present an extensive list of features used in
\langid, some of which are not self-evident. The equations written in
the unified notation defined earlier show how the values \(v\) used in
the language models are calculated from the tokens \(u\). For each
feature type, we generally introduce the first published article that
used that feature type, as well as more recent articles where the
feature type has been considered.

\subsection{Bytes and Encodings}

In \langid, text is typically modeled as a stream of
characters. However, there is a slight mismatch between this view and
how text is actually stored: documents are digitized using a particular
encoding, which is a mapping from characters (e.g.\ a character in an
alphabet), onto the actual sequence of bytes that is stored and
transmitted by computers. Encodings vary in how many bytes they use to
represent each character. Some encodings use a fixed number of bytes for
each character (e.g.\ ASCII), whereas others use a variable-length
encoding (e.g.\ UTF-8). Some encodings are specific to a given language
(e.g.\ GuoBiao 18030 or Big5 for Chinese), whereas others are
specifically designed to represent as many languages as possible (e.g.\
the Unicode family of encodings). Languages can often be represented in
a number of different encodings (e.g.\ UTF-8 and Shift-JIS for
Japanese), and sometimes encodings are specifically designed to share
certain codepoints (e.g.\ all single-byte UTF-8 codepoints are exactly
the same as ASCII). Most troubling for \langid, isomorphic encodings can
be used to encode different languages, meaning that the determination of
the encoding often doesn't help in honing in on the language. Infamous
examples of this are the ISO-8859 and EUC encoding families.  Encodings
pose unique challenges for practical \langid applications: a given
language can often be encoded in different forms, and a given encoding
can often map onto multiple languages.

Some \langid research has included an explicit encoding detection step
to resolve bytes to the characters they represent \citep{kikui1},
effectively transcoding the document into a standardized encoding before
attempting to identify the language.  However, transcoding is
computationally expensive, and other research suggests that it may be
possible to ignore encoding and build a single per-language model
covering multiple encodings simultaneously
\citep{kruengkrai2,baldwin2}. Another solution is to treat each
language-encoding pair as a separate category
\citep{Cowie+:1999,suzuki1,singh2,brown1}. The disadvantage of this is
that it increases the computational cost by modeling a larger number of
classes.  Most of the research has avoided issues of encoding entirely
by assuming that all documents use the same encoding \citep{mandl1}. This
may be a reasonable assumption in some settings, such as when processing
data from a single source (e.g.\ all data from Twitter and Wikipedia is
UTF-8 encoded). In practice, a disadvantage of this approach may be that
some encodings are only applicable to certain languages (e.g.\ S-JIS for
Japanese and Big5 for Chinese), so knowing that a document is in a
particular encoding can provide information that would be lost if the
document is transcoded to a universal encoding such as
UTF-8. \cite{li2} used a parallel state machine to detect which
encoding scheme a file could potentially have been encoded with. The
knowledge of the encoding, if detected, is then used to narrow down the
possible languages.

Most features and methods do not make a distinction between bytes or
characters, and because of this we will present feature and method
descriptions in terms of characters, even if byte tokenization was
actually used in the original research.

\subsection{Characters}
\label{sec:characters}

In this section, we review how individual character tokens have been used as features in \langid.

\paragraph{Non-alphabetic or non-ideographic characters}

\cite{bali2} used the formatting of numbers when distinguishing between
Malay and Indonesian. \cite{king1} used the presence of non-alphabetic
characters between the current word and the words before and after as
features. \cite{elfardy2} used emoticons (or emojis) in Arabic dialect
identification with Naive Bayes (``NB''; see
\secref{product}). Non-alphabetic characters have also been used by
\cite{basile1}, \cite{bestgen1}, \cite{samih3}, and \cite{simaki1}.

\paragraph{Alphabets}

\cite{henrich1} used knowledge of alphabets to exclude languages where
a language-unique character in a test document did not
appear. \cite{giguet2} used alphabets collected from dictionaries to
check if a word might belong to a language. \cite{hanif1} used the
Unicode database to get the possible languages of individual Unicode
characters. Lately, the knowledge of relevant alphabets has been used
for \langid also by \cite{hasimu1} and \cite{samih3}.

\paragraph{Capitalization}

Capitalization is mostly preserved when calculating character \ngram
frequencies, but in contexts where it is possible to identify the
orthography of a given document and where capitalization exists in the
orthography, lowercasing can be used to reduce sparseness. In recent
\langid work, capitalization was used as a special feature by
\cite{basile1}, \cite{bestgen1}, and \cite{simaki1}.

\paragraph{The number of characters in words and word combinations}

\cite{langer1} was the first to use the length of words in \langid. \cite{nobesawa1} used the length of full person names comprising several words. Lately, the number of characters in words has been used for \langid by \cite{dongen1}, \cite{lee3}, \cite{samih3}, and \cite{simaki1}. \cite{dongen1} also used the length of the two preceding words.

\paragraph{The frequency or probability of each character}

\cite{kerwin1} used character frequencies as feature vectors. In a
feature vector, each feature \(f\) has its own integer value. The raw
frequency -- also called term frequency (TF) -- is calculated for each
language \(g\) as:
\begin{equation}
v_{C_g}(f)= c(C_g,f)
\label{eqn:frequency}
\end{equation}
\cite{rau1} was the first to use the probability of characters. He
calculated the probabilities as relative frequencies, by dividing the frequency of a feature found
in the corpus by the total count of features of the same type in the
corpus. When the relative frequency of a feature \(f\) is used as a
value, it is calculated for each language \(g\) as:

\begin{equation}
v_{C_g}(f)=\frac {c(C_g,f)} {l_{C_g^{F}}}
\label{eqn:relativefrequency}
\end{equation}
\cite{tran1} calculated the relative frequencies of one character prefixes, and \cite{windisch1} did the same for one character suffixes. 

\cite{ng3} calculated character frequency document frequency (``LFDF'')
values. \cite{takci2} compared their own Inverse Class Frequency (``ICF'')
method with the Arithmetic Average Centroid (``AAC'') and the Class Feature
Centroid (``CFC'') feature vector updating methods. In ICF a character
appearing frequently only in some language gets more positive weight for
that language. The values differ from Inverse Document Frequency (``IDF'',
\eqnref{artemenko1}), as they are calculated using also the
frequencies of characters in other languages. Their ICF-based vectors
generally performed better than those based on AAC or
CFC. \cite{takci5} explored using the relative frequencies of
characters with similar discriminating weights. \cite{takci2} also used
Mutual Information (``MI'') and chi-square weighting schemes with
characters.

\cite{baldwin2} compared the identification results of single
characters with the use of character bigrams and trigrams when
classifying over 67 languages. Both bigrams and trigrams generally
performed better than unigrams. \cite{jauhiainen1} also found that the
identification results from identifiers using just characters are
generally worse than those using character sequences.

\subsection{Character Combinations}

In this section we consider the different combinations of characters used in the literature. Character \ngrams mostly consist of all possible characters in a given encoding, but can also consist of only alphabetic or ideographic characters.

\paragraph{Co-occurrences}

\cite{windisch1} calculated the co-occurrence ratios of any two characters, as well as the ratio of consonant clusters of different sizes to the total number of consonants. \cite{sterneberg1} used the combination of every bigram and their counts in words.  \cite{lee3} used the proportions of question and exclamation marks to the total number of the end of sentence punctuation as features with several machine learning algorithms.

\cite{francosalvador4} used FastText to generate character \emph{n}-gram embeddings \citep{joulin1}. Neural network generated embeddings are explained in \secref{cooccurrencesofwords}.

\paragraph{Vowel-consonant relationship}

\cite{rau1} used the relative frequencies of vowels following vowels, consonants following vowels, vowels following consonants and consonants following consonants. \cite{dongen1} used vowel-consonant ratios as one of the features with Support Vector Machines (``SVMs'', \secref{supportvectormachines}), Decision Trees (``DTs'', \secref{decisiontrees}), and Conditional Random Fields (``CRFs'', \secref{openissues:short}). 

\paragraph{Character repetition}

\cite{elfardy2} used the existence of word lengthening effects and
repeated punctuation as features. \cite{banerjee1} used the presence of
characters repeating more than twice in a row as a feature with simple
scoring (\eqnref{simple1}). \cite{barman1} used more complicated
repetitions identified by regular expressions. \cite{sikdar1} used
letter and character bigram repetition with a CRF. \cite{martinc1} used
the count of character sequences with three or more identical
characters, using several machine learning algorithms.

\paragraph{\ngrams of characters of the same size}

Character \ngrams are continuous sequences of characters of length
$n$. They can be either consecutive or overlapping. Consecutive
character bigrams created from the four character sequence \lex{door}
are \lex{do} and \lex{or}, whereas the overlapping bigrams are \lex{do},
\lex{oo}, and \lex{or}. Overlapping \ngrams are most often used in the
literature. Overlapping produces a greater number and variety of \ngrams
from the same amount of text.

\cite{rau1} was the first to use combinations of any two characters. He
calculated the relative frequency of each bigram. \tabref{RFTable2}
lists more recent articles where relative frequencies of \ngrams of
characters have been used. \cite{rau1} also used the relative
frequencies of two character combinations which had one unknown
character between them, also known as gapped bigrams. \cite{seifart1} used a
modified relative frequency of character unigrams and bigrams.

\begin{table}[t!]
\footnotesize
\centering
\begin{tabular}{p{8cm}ccccccc}
\hline
\textbf{Article} & \textbf{1} & \textbf{2} & \textbf{3}	 & \textbf{4} & \textbf{5} & \textbf{6} & \textbf{7} \\
\hline
\cite{bosnjak1} &  &  &  & \bestone& & & \\
\cite{zampieri3} &  & \bestother & \besttwo & \bestone & \bestother & & \\
\cite{das1} & \bestother & \bestother & \besttwo & \besttwo & \bestother & \bestother & \bestother \\
\cite{indhuja2} & \bestone & &  & & & & \\
\cite{ljubesic4} & & &\bestOther& & &\bestOther& \\
\cite{minocha1} &\bestother&\bestother& \bestone &\bestother&\bestother& & \\
\cite{petho1} &  &\bestother&\bestother& \besttwo & \bestone & & \\
\cite{sadat2,sadat1} &\bestother& \bestone &\bestother& & & & \\
\cite{tan3} & &\bestother&\bestother&\bestother& \bestone & \besttwo & \\
\cite{zaidan1} &\bestother& & \besttwo & & \bestone & & \\
\cite{zampieri5} & & & \bestone & & & & \\
\cite{francosalvador2} & & & & \bestone & & & \\
\cite{jauhiainen2} &\bestother&\bestother&\bestother&\bestother&\bestother& \bestone & \besttwo ... \\
\cite{king4} & & &   & & \bestone& & \\
\cite{panich1} &\bestOther&\bestOther&\bestOther&\bestOther&\bestOther&\bestOther& \\
\cite{zampieri7} & & & & & \bestone & & \\
\cite{abainia2} &\bestOther&\bestOther&\bestOther& & & & \\
\cite{castro1,castro2} & &\bestother&\bestother&\bestother&\bestother& \bestone & \besttwo \\
\cite{giwa3} &\bestother&\bestother& \besttwo & \bestone &\bestother&\bestother&\bestother\\
\cite{hanani1} &  &  & \bestone  & & & & \\
\cite{duvenhage1} & & &   & & \bestone & & \\
\cite{jourlin1} & \bestone \\
\hline
\end{tabular}
\caption{List of articles (2013--2017) where relative frequencies of
  character \ngrams have been used as features. The columns indicate the
  length of the \ngrams used. ``\bestone'' indicates the empirically
  best \ngram length in that paper, and ``\besttwo'' the second-best
  \ngram length. ``\bestother'' indicates that there was no clear winner
  in terms of \ngram order reported on in the paper.}
\label{tab:RFTable2}
\end{table}

Character trigram frequencies relative to the word count were used by
\cite{vega1}, who calculated the values \(v_{C}(f)\) as in
\eqnref{vega1}. Let \(T\) be the word-tokenized segmentation of the
corpus \(C\) of character tokens, then:
\begin{equation}
\label{eqn:vega1}
v_{C}(f) =  \frac{c(C,f)}{l_T}
\end{equation}
where \(c(C,f)\) is the count of character trigrams \(f\) in $C$, and
\(l_T\) is the total word count in the corpus. Later \ngram frequencies relative to the word count were used by \cite{hamzah1} for character bigrams and trigrams.

\cite{house1} divided characters into five phonetic groups and used a
Markovian method to calculate the probability of each bigram consisting
of these phonetic groups. In Markovian methods, the probability of a
given character \(u_i\) is calculated relative to a fixed-size character
context \(u_{i-n+1}, ..., u_{i-1}\) in corpus
\(C\), as follows:
\begin{equation}
\label{eqn:rmarkovianprobability}
P(u_{i}|u_{i-n+1, ..., i-1})=\frac {c(C,u_{i-n+1, ..., i})} {c(C,u_{i-n+1, ..., i-1})}
\end{equation}
where \(u_{i-n+1, ..., i-1}\) is an \ngram prefix of
\(u_{i-n+1, ..., i}\) of length \(n-1\) . In this case, the probability
\(P(u_{i}|u_{i-n+1, ..., i-1})\) is the value \(v_{C}(f)\), where
\(f = u_{i-n+1, ..., i}\), in the model \(O(C)\). \cite{ludovik1} used
4-grams with recognition weights which were derived from Markovian
probabilities. \tabref{MarkovianTable} lists some of the more recent
articles where Markovian character \ngrams have been used.

\begin{table}[t!]
\centering
\footnotesize
\begin{tabular}{p{8.5cm}cccccccc}
\hline
\textbf{Article} & \textbf{2} & \textbf{3}	 & \textbf{4} & \textbf{5} & \textbf{6} & \textbf{7} & \textbf{8} \\
\hline
\cite{ramisch1} & \besttwo & \besttwo & \besttwo & \bestone\\
\cite{you1} & & \besttwo & \bestone & \besttwo &\bestother& \bestother\\
\cite{stupar1,tiedemann1} & & \bestone\\
\cite{goldszmidt1} & & \bestone\\
\cite{bar1} & & \bestone & & & & &\\
\cite{brown3} & & \bestone\\
\cite{gamallo1}  & & \bestone & & & & & &\\
\cite{hurtado1} & & & \bestone\\
\cite{indhuja2} &\bestOther& \bestOther\\
\cite{leidig2} & & & & \bestone & & &\\
\cite{mendizabal1} &\bestother&\bestother& \besttwo & \bestone & \besttwo & \besttwo & \besttwo\\
\cite{petho1} &\bestother&\bestother& \besttwo & \bestone\\
\cite{sadat2,sadat1} & \bestone & \besttwo & & & \\
\cite{ullman1} & & & \bestone\\
\cite{cianflone1} &\bestother&\bestother&\bestother&\bestother&\bestother& \besttwo & \bestone \\
\cite{martadinata1} & &\bestother& \besttwo & \bestone &\\
\cite{samih2,samih3} & & & & \bestone & & & \\
\hline
\end{tabular}
\caption{List of recent articles where Markovian character \ngrams have
  been used as features. The columns indicate the length of the \ngrams
  used. ``\bestone'' indicates the best and ``\besttwo'' the second-best
  \ngram length as reported in the article in question. ``\bestother'' indicates that there was no clear winner
  in terms of \ngram order reported on in the paper.}
\label{tab:MarkovianTable}
\end{table}

\cite{vitale1} was the first author to propose a full-fledged
probabilistic language identifier. He defines the probability of a
trigram \(f\) being written in the language \(g\) to be:
\begin{equation}
\label{eqn:vitale1}
P(g|f) = \frac {P(f|g)P(g)} {\sum_{h \in G}P(f|h)P(h)}
\end{equation}
He considers the prior probabilities of each language \(P(g)\) to be equal, which leads to:
\begin{equation}
\label{eqn:vitale2}
P(g|f) = \frac {P(f|g)} {\sum_{h \in G}P(f|h)}
\end{equation}
\cite{vitale1} used the probabilities \(P(g|f)\) as the values \(v_{C_{g}}(f)\) in the language models.

\cite{macnamara1} used a list of the most frequent bigrams and trigrams
with logarithmic weighting. \cite{prager1} was the first to use direct
frequencies of character \ngrams as feature vectors. \cite{vinosh1}
used Principal Component Analysis (``PCA'') to select only the most
discriminating bigrams in the feature vectors representing
languages. \cite{murthy1} used the most frequent and discriminating
byte unigrams, bigrams, and trigrams among  their feature functions. They define
the most discriminating features as those which have the most differing
relative frequencies between the models of the different
languages. \cite{gottron1} tested \ngrams from two to five using
frequencies as feature vectors, frequency ordered lists, relative
frequencies, and Markovian probabilities. \tabref{FrequencyVectorTable}
lists the more recent articles where the frequency of character \ngrams
have been used as features. In the method column, ``RF'' refers to Random
Forest (cf.\ \secref{decisiontrees}), ``LR'' to Logistic Regression
(\secref{discriminantfunctions}), ``KRR'' to Kernel Ridge Regression
(\secref{vectors}), ``KDA'' to Kernel Discriminant Analysis
(\secref{vectors}), and ``NN'' to Neural Networks (\secref{neuralnetworks}).

\begin{table}[t]
\footnotesize
\centering
\begin{tabular}{p{6cm}ccccccp{4.4cm}}
\hline
\textbf{Article} & \textbf{1} & \textbf{2}	 & \textbf{3} & \textbf{4} & \textbf{5} & \textbf{6+} & \textbf{Method} \\
\hline
\cite{adouane2} &\bestOther&\bestOther&\bestOther&\bestOther&\bestOther&\bestOther& SVM (cf.\ \ssecref{supportvectormachines})\\
\cite{albadrashiny3} &\bestOther&\bestOther&\bestOther&\bestOther&\bestOther& & CRF (cf.\ \ssecref{openissues:short})\\
\cite{alshutayri1} &\bestOther&\bestOther&\bestOther& & & & SVM, DT (cf.\ \ssecref{decisiontrees})\\
\cite{barbaresi1} & &\bestOther&\bestOther&\bestOther&\bestOther&\bestOther & NB (cf.\ \ssecref{product}), XGBoost (cf.\ \ssecref{othermethods}), RF (cf.\ \ssecref{decisiontrees})\\
\cite{ciobanu2} &\bestOther&\bestOther&\bestOther&\bestOther& & & LR (cf.\ \ssecref{discriminantfunctions})\\
\cite{giwa3} & &\bestOther&\bestOther&\bestOther&\bestOther& & SVM\\
\cite{goutte4} & & & & & &\bestOther& SVM, NB\\
\cite{ionescu1} & &\bestOther&\bestOther&\bestOther&\bestOther&\bestOther & KRR (cf.\ \ssecref{vectors}), KDA (cf.\ \ssecref{vectors})\\
\cite{lamabam1} & & &\bestOther& & & & CRF\\
\cite{criscuolo1} & & & & &\bestOther& & NB \\
\cite{malmasi5} &\bestOther&\bestOther&\bestOther&\bestOther&\bestOther&\bestOther& SVM \\
\cite{malmasi6} & & & &\bestOther& & & SVM \\
\cite{mathur1} &\bestOther&\bestOther&\bestOther&\bestOther&\bestOther&\bestOther& NB, LR \\
\cite{oliveira1} & & &\bestOther&\bestOther&\bestOther&\bestOther& SVM \\
\cite{cagigos1} & & &\bestOther& & & & NB, SVM, DT, NN (cf.\ \ssecref{neuralnetworks})\\
\cite{rangel1} & & & &\bestOther& & & SVM, NB, NN, ... \\
\cite{schaetti1} & &\bestOther& & & & & NN   \\
\hline
\end{tabular}
\caption{Recent papers (2016--) where the frequency of character \ngrams
  has been used to generate feature vectors. The columns indicate the
  length of the \ngrams used, and the machine learning method(s) used. The relevant
  section numbers for the methods are mentioned in parentheses.}
\label{tab:FrequencyVectorTable}
\end{table}

\cite{giguet2} used the last two and three characters of
open class words.
\cite{suzuki1} used an unordered list of distinct trigrams with the
simple scoring method (\secref{Simplescoring}). \cite{hayati1} used
Fisher's discriminant function to choose the 1000 most discriminating
trigrams. \cite{bilcu1} used unique 4-grams of characters with positive Decision Rules (\secref{Decisionrule}).
\cite{ozbek1} used the frequencies of bi- and
trigrams in words unique to a language. \cite{milne1} used lists of the
most frequent trigrams.

\cite{li2} divided possible character bigrams into those that are
commonly used in a language and to those that are not. They used the
ratio of the commonly used bigrams to all observed bigrams to give a confidence score for each language.
\cite{xafopoulos1} used the difference between the ISO
Latin-1 code values of two consecutive characters as well as two
characters separated by another character, also known as gapped
character bigrams.

\cite{artemenko2} used the IDF and the transition probability of trigrams. They calculated the IDF values \(v_{C_{g}}(f)\) of trigrams \(f\) for each language \(g\), as in \eqnref{artemenko1}, where \(c(C_{g},f)\) is the number of trigrams \(f\) in the corpus of the language \(g\) and \(df(C_{G},f)\) is the number of languages in which the trigram \(f\) is found, where \(C_{G}\) is the language-segmented training corpus with each language as a single segment.

\begin{equation}
v_{C_{g}}(f)=\frac {c(C_{g},f)}{df(C_{G},f)}
\label{eqn:artemenko1}
\end{equation}
\(df\) is defined as:
\begin{equation}
df(C_{G},f) = \sum_{g\in G} \left\{ 
  \begin{array}{l l}
1 & \quad \text{, if }c(C_{g},f)>0\\
0 & \quad \text{, otherwise}
  \end{array} \right.
\label{eqn:df}
\end{equation}
\cite{malmasi3} used \ngrams from one to four, which were weighted with ``TF-IDF'' (Term Frequency--Inverse Document Frequency). TF-IDF was calculated as:
\begin{equation}
\label{eqn:tfidf}
v_{C_{g}}(f)=c(C_g,f) \log \frac {l_{G}}{df(C_{G},f)}
\end{equation}
TF-IDF weighting or close variants have been widely used for \langid. \cite{thomas1} used ``CF-IOF'' (Class Frequency-Inverse Overall Frequency) weighted 3- and 4-grams.

\cite{jhamtani1} used the logarithm of the ratio of the counts of
character bigrams and trigrams in the English and Hindi
dictionaries. \cite{zamora1} used a feature weighting scheme based on
mutual information (``MI''). They also tried weighting schemes based on
the ``GSS'' (Galavotti, Sebastiani, and Simi) and ``NGL'' (Ng, Goh, and Low)
coefficients, but using the MI-based weighting scheme proved the best in
their evaluations when they used the sum of values method
(\eqnref{sumvalues1}). \cite{martinc1} used punctuation trigrams, where
the first character has to be a punctuation mark (but not the other two
characters). \cite{saharia1} used consonant bi- and trigrams which were
generated from words after the vowels had been removed.

\paragraph{Character \ngrams of differing sizes}

The language models mentioned earlier consisted only of \ngrams of the
same size \(n\). If \ngrams from one to four were used, then there were
four separate language models. \cite{cavnar1} created ordered lists of
the most frequent \ngrams for each language. \cite{singh4} used similar
\ngram lists with symmetric cross-entropy. \cite{russell1} used a
Markovian method to calculate the probability of byte trigrams
interpolated with byte unigrams. \cite{vatanen1} created a language
identifier based on character \ngrams of different sizes over 281
languages, and obtained an identification accuracy of 62.8\% for
extremely short samples (5--9 characters). Their language identifier was
used or evaluated by \cite{rodrigues1}, \cite{maier1}, and
\cite{jauhiainen6}. \cite{rodrigues1} managed to improve the
identification results by feeding the raw language distance calculations
into an SVM.

\tabref{DifferingNgramTable3} lists recent articles where character
\ngrams of differing sizes have been used. ``LR'' in the methods column
refer to Logistic Regression (\secref{maxent}), ``LSTM RNN'' to Long
Short-Term Memory Recurrent Neural Networks (\secref{neuralnetworks}),
and ``DAN'' to Deep Averaging Networks
(\secref{neuralnetworks}). \cite{kikui1} used up to the four last
characters of words and calculated their relative
frequencies. \cite{ahmed1} used frequencies of 2--7-grams, normalized
relative to the total number of \ngrams in all the language models as
well as the current language model. \cite{jauhiainen1} compared the use
of different sizes of \ngrams in differing combinations, and found that
combining \ngrams of differing sizes resulted in better identification
scores. \cite{lui1,lui2,lui5} used mixed length domain-independent
language models of byte \ngrams from one to three or four.

Mixed length language models were also generated by \cite{brown1} and
later by \cite{brown2,brown3}, who used the most frequent and
discriminating \ngrams longer than two bytes, up to a maximum of 12
bytes, based on their weighted relative frequencies. \(K\) of the most
frequent \ngrams were extracted from training corpora for each language,
and their relative frequencies were calculated. In the tests reported in
\citep{brown2}, \(K\) varied from 200 to 3,500 \ngrams. Later 
\cite{sanchezperez1} also evaluated different combinations of character
\ngrams as well as their combinations with words.

\cite{stensby1} used mixed-order \ngram frequencies relative to the
total number of \ngrams in the language model. \cite{sterneberg1} used
frequencies of \ngrams from one to five and gapped 3- and 4-grams as
features with an SVM. As an example, some gapped 4-grams from the word
\lex{Sterneberg} would be \lex{Senb}, \lex{tree}, \lex{enbr}, and
\lex{reeg}. \cite{king5} used character \ngrams as a backoff from
Markovian word \ngrams. \cite{shrestha1} used the frequencies of word
initial \ngrams ranging from 3 to the length of the word minus
1. \cite{acs1} used the most relevant \ngrams selected using the
absolute value of the Pearson correlation. \cite{mandal1} used only the
first 10 characters from a longer word to generate the \ngrams, while
the rest were ignored. \cite{qiao1} used only those \ngrams which had
the highest TF-IDF scores. \cite{bestgen1} used character \ngrams
weighted by means of the ``BM25'' (Best Match 25) weighting
scheme. \cite{hanani2} used byte \ngrams up to length 25.

\begin{table}[h]
\footnotesize
\begin{tabular}{p{6cm}llllllllp{3cm}}
\hline
\textbf{Article} & \textbf{1} & \textbf{2} & \textbf{3}	 & \textbf{4} & \textbf{5} & \textbf{6} & \textbf{7} & \textbf{8+} & \textbf{Method}\\
\hline
\cite{adouane3,adouane1} &\bestOther&\bestOther&\bestOther& & & & & & SVM (cf.\ \ssecref{supportvectormachines})\\
\cite{adouane5,adouane4} & & & & &\bestOther&\bestOther& & & SVM \\
\cite{balazevic1} &\bestOther&\bestOther&\bestOther&\bestOther& & & & & Product (cf.\ \ssecref{product}), SVM, LR (cf.\ \ssecref{discriminantfunctions}), Sum (cf.\ \ssecref{sum})\\
\cite{coltekin1} &\bestOther&\bestOther&\bestOther&\bestOther&\bestOther&\bestOther& & & SVM \\ 
\cite{eldesouki1} & &\bestOther&\bestOther&\bestOther&\bestOther& & & & SVM, LR, NN (cf.\ \ssecref{neuralnetworks}), NB (cf.\ \ssecref{product})\\ 
\cite{he1} &\bestOther&\bestOther&\bestOther&\bestOther& & & & & LR (cf.\ \ssecref{maxent})\\
\cite{jauhiainen5} &\bestOther&\bestOther&\bestOther&\bestOther&\bestOther&\bestOther&\bestOther&\bestOther& HeLI (cf.\ \ssecref{product})\\
\cite{malmasi4,malmasi7} &\bestOther&\bestOther&\bestOther&\bestOther&\bestOther&\bestOther& & & SVM\\
\cite{piergallini2} &\bestOther&\bestOther&\bestOther&\bestOther& & & & & LR\\
\cite{piergallini1} &\bestOther&\bestOther&\bestOther& & & & & & LR\\
\cite{radford1} & &\bestOther&\bestOther&\bestOther&\bestOther& & & & LR\\
\cite{samih1} & &\bestOther&\bestOther& & & & & & LSTM RNN (cf.\ \ssecref{neuralnetworks})\\
\cite{xu1} &\bestOther&\bestOther&\bestOther& & & & & & SVM \\
\cite{alrifai1} & &\bestOther&\bestOther&\bestOther&\bestOther&\bestOther&\bestOther& & SVM \\
\cite{barbaresi2} & &\bestOther&\bestOther&\bestOther&\bestOther&\bestOther&\bestOther& & NB \\
\cite{bestgen1} &\bestOther&\bestOther&\bestOther&\bestOther&\bestOther&\bestOther&\bestOther&\bestOther& SVM \\
\cite{clematide1} &\bestOther&\bestOther&\bestOther&\bestOther&\bestOther&\bestOther& & & NB, CRF (cf.\ \ssecref{openissues:short}), SVM \\
\cite{linares1} & &\bestOther&\bestOther& & & & & & SVM, NB, ... \\
\cite{francosalvador4} & & &\bestOther&\bestOther&\bestOther&\bestOther& & & DAN (cf.\ \ssecref{neuralnetworks})\\
\cite{gamallo3} & & & & &\bestOther&\bestOther&\bestOther& & Perplexity (cf.\ \ssecref{perplexity})\\
\cite{gomezadorno1} & & &\bestOther&\bestOther&\bestOther& & & & NB, SVM\\
\cite{hanani2} &\bestOther&\bestOther&\bestOther& & & & & & SVM, NB, LR, DT (cf.\ \ssecref{decisiontrees}) \\
\cite{jauhiainen7} &\bestOther&\bestOther&\bestOther&\bestOther&\bestOther&\bestOther&\bestOther&\bestOther& HeLI\\
\cite{jauhiainen6} &\bestOther&\bestOther&\bestOther&\bestOther&\bestOther&\bestOther& & & HeLI\\
\cite{malmasi5} &\bestOther&\bestOther&\bestOther&\bestOther&\bestOther&\bestOther& & & SVM\\
\cite{malmasi8} &\bestOther&\bestOther&\bestOther&\bestOther&\bestOther&\bestOther&\bestOther&\bestOther& SVM\\
\cite{mathur1} & &\bestOther&\bestOther&\bestOther&\bestOther& & & & RNN\\
\cite{miura1} &\bestOther&\bestOther&\bestOther&\bestOther&\bestOther&\bestOther& & & SVM \\
\cite{sanchezperez1} & & &\bestOther&\bestOther&\bestOther&\bestOther&\bestOther&\bestOther& SVM\\
\cite{tellez1} &\bestOther& &\bestOther& &\bestOther& &\bestOther&\bestOther & SVM\\
\cite{linares2} & &\bestOther&\bestOther&\bestOther& & & & & SVM, NB, ...  \\
\hline
\end{tabular}
\caption{\small{List of articles (2016-) where character \ngrams of differing sizes have been used as features. The numbered columns indicate the length of the \ngrams used. The method column indicates the method used with the \ngrams. The relevant section numbers are mentioned in parentheses.}}
\label{tab:DifferingNgramTable3}
\end{table}

\paragraph{Consonant or vowel sequences}

\cite{sterneberg1} used consonant sequences generated from words. \cite{anand1} used the presence of vowel sequences as a feature with a NB classifier (see \secref{naivebayes}) when distinguishing between English and transliterated Indian languages.

\paragraph{\ngram dictionary}

\cite{chanda1} used a basic dictionary (\secref{basicdictionary}) composed of the 400 most common character 4-grams.

\paragraph{Unique character combinations}

\cite{henrich1} and \cite{vitale1} used character combinations (of different sizes) that either existed in only one language or did not exist in one or more languages.

\subsection{Morphemes, Syllables and Chunks}

\paragraph{Morphemes}

\cite{giguet3} used the suffixes of lexical words derived from untagged
corpora. \cite{el-shishiny1} used prefixes and suffixes determined
using linguistic knowledge of the Arabic language. \cite{marcadet1}
used suffixes and prefixes in rule-based \langid. \cite{ozbek1} used
morphemes and morpheme trigrams (morphotactics) constructed by Creutz's
algorithm \citep{creutz1}. \cite{hammarstrom1} used prefixes and
suffixes constructed by his own algorithm, which was later also used by
\cite{ceylan1}. \cite{romsdorfer2} used morpheme lexicons in
\langid. \cite{ceylan1} compared the use of morphological features with
the use of variable sized character \ngrams. When choosing between ten
European languages, the morphological features obtained only 26.0\%
accuracy while the \ngrams reached 82.7\%. \cite{yeong1} lemmatized Malay
words in order to get the base forms. \cite{lu1} used a
morphological analyzer of Arabic. \cite{zampieri3} used morphological
information from a part-of-speech (POS) tagger. \cite{anand1} and
\cite{banerjee1} used manually selected suffixes as features.
\cite{becavac1} created morphological grammars to distinguish between
Croatian and Serbian. \cite{darwish1} used morphemes created by
Morfessor, but they also used manually created morphological
rules. \cite{gamallo1} used a suffix module containing the most
frequent suffixes. \cite{dutta1} and \cite{mandal1} used word suffixes
as features with CRFs. \cite{barbaresi1} used an unsupervised method to
learn morphological features from training data. The method collects
candidate affixes from a dictionary built using the training data. If
the remaining part of a word is found from the dictionary after removing
a candidate affix, the candidate affix is considered to be a morpheme. \cite{barbaresi1} used 5\% of the most frequent affixes in language identification.
\cite{gomezadorno1} used character \ngrams classified into different types, which included prefixes and suffixes.
\tabref{PrefixSuffixTable} lists some of the more recent articles where
prefixes and suffixes collected from a training corpus has been used for
\langid.

\begin{table}[t!]
\footnotesize
\begin{tabular}{llllll}
\hline
\textbf{Reference} & \textbf{1} & \textbf{2} & \textbf{3} & \textbf{4} & \textbf{Method}\\
\hline
\cite{he1} &\bestOther&\bestOther&\bestOther& & LR (cf.\ \ssecref{maxent})\\  
\cite{piergallini2} &\bestOther&\bestOther&\bestOther&\bestOther& LR\\
\cite{samih2,samih3} &\bestOther&\bestOther&\bestOther& & CRF (cf.\ \ssecref{openissues:short})\\  
\cite{schulz1} &\bestOther&\bestOther&\bestOther& & CRF\\  
\cite{shrestha2} &\bestOther&\bestOther&\bestOther&\bestOther& CRF\\  
\cite{sikdar1} &\bestOther&\bestOther&\bestOther&\bestOther& CRF\\
\cite{xia5} &\bestOther&\bestOther&\bestOther& & CRF\\  
\cite{clematide1} &\bestOther&\bestOther&\bestOther& & CRF\\  
\cite{gomezadorno1} & & &\bestOther& & NB (cf.\ \ssecref{product}), SVM (cf.\ \ssecref{supportvectormachines})\\
\cite{martinc1} & & & &\bestOther& SVM, LR, RF (cf.\ \ssecref{decisiontrees}), ...  \\
\hline
\end{tabular}
\caption{References (2016-) where prefixes and suffixes collected from a training corpus have been used for \langid. The columns indicate the length of the prefixes and suffixes. The method column indicates the method used. The relevant section numbers are mentioned in parentheses.}
\label{tab:PrefixSuffixTable}
\end{table}

\paragraph{Syllables and syllable \ngrams}

\cite{chen2} used trigrams composed of syllables. \cite{yeong1} used
Markovian syllable bigrams for \langid between Malay and English. Later
\cite{yeong2} also experimented with syllable uni- and
trigrams. \cite{murthy1} used the most frequent as well as the most
discriminating Indian script syllables, called aksharas. They used
single aksharas, akshara bigrams, and akshara trigrams. Syllables would
seem to be especially apt in situations where distinction needs to be
made between two closely-related languages.

\paragraph{Chunks, chunk \ngrams and \ngrams of \ngrams}

\cite{you1} used the trigrams of non-syllable chunks that were based on
MI. \cite{yeong1} experimented also with Markovian bigrams using both
character and grapheme bigrams, but the syllable bigrams proved to work
better. Graphemes in this case are the minimal units of the writing
system, where a single character may be composed of several graphemes
(e.g.\ in the case of the Hangul or Thai writing systems). Later,
\cite{yeong2} also used grapheme uni- and trigrams. \cite{yeong2}
achieved their best results combining word unigrams and syllable bigrams
with a grapheme back-off. \cite{elfardy4} used the MADAMIRA toolkit for
D3 decliticization and then used D3-token 5-grams. D3 decliticization is
a way to preprocess Arabic words presented by \cite{habash1}.

Graphones are sequences of characters linked to sequences of
corresponding phonemes. They are automatically deduced from a bilingual
corpus which consists of words and their correct pronunciations using
Joint Sequence Models (``JSM''). \cite{giwa2} used language tags instead of phonemes when generating the graphones and then used Markovian graphone
\ngrams from 1 to 8 in \langid.

\subsection{Words}

\paragraph{Position of words}

\cite{kumar1} used the position of the current word in word-level
\langid. The position of words in sentences has also been used as a
feature in code-switching detection by \cite{dongen1}. It had
predictive power greater than the language label or length of the previous word.

\paragraph{The characteristics of words}

\cite{mustonen1} used the characteristics of words as parts of discriminating functions. \cite{barman2} used the string edit distance and \ngram overlap between the word to be identified and words in dictionaries. Similarly \cite{jhamtani1} used a modified edit distance, which considers the common spelling substitutions when Hindi is written using latin characters. \cite{das2} used the Minimum Edit Distance (``MED''). 

\paragraph{Basic dictionary}
\label{sec:basicdictionary}

Basic dictionaries are unordered lists of words belonging to a
language. Basic dictionaries do not include information about word
frequency, and are independent of the dictionaries of other
languages. \cite{vitale1} used a dictionary for \langid as a part of
his speech synthesizer. Each word in a dictionary had only one possible
``language'', or pronunciation category. More recently, a basic
dictionary has been used for \langid by \cite{adouane6},
\cite{dongen1}, and \cite{duvenhage1}.

\paragraph{Dictionary of unique words}

Unique word dictionaries include only those words of the language, that
do not belong to the other languages targeted by the language
identifier. \cite{kulikowski1} used unique short words (from one to
three characters) to differentiate between languages. Recently, a
dictionary of unique words was used for \langid by
\cite{adouane2}, \cite{guellil1}, and \cite{martinc1}.

\paragraph{Specific classes of words}

\cite{giguet2} used exhaustive lists of function words collected
from dictionaries.
\cite{wechsler1} used stop words -- that is non-content or closed-class
words -- as a training corpus. Similarly, \cite{lins1}
used words from closed word classes, and \cite{stupar1} used lists of
function words.  \cite{albadrashiny1} used a lexicon of Arabic words
and phrases that convey modality. Common to these features is that they
are determined based on linguistic knowledge.

\paragraph{Discriminating words}

\cite{rehurek1} used the most relevant words for each
language. \cite{babu1} used unique or nearly unique
words. \cite{francosalvador2} used Information Gain Word-Patterns
(``IG-WP'') to select the words with the highest information gain.

\paragraph{Most common words}

\cite{souter1} made an (unordered) list of the most common words for
each language, as, more recently, did \cite{cazamias1},
\cite{panich1}, and \cite{abainia2}. \cite{pavan1} encoded the most
common words to root forms with the Soundex algorithm.

\paragraph{Word frequency}

\cite{mather1} collected the frequencies of words into feature
vectors. \cite{prager1} compared the use of character \ngrams from 2 to
5 with the use of words. Using words resulted in better identification
results than using character bigrams (test document sizes of 20, 50, 100
or 200 characters), but always worse than character 3-, 4- or
5-grams. However, the combined use of words and character 4-grams gave
the best results of all tested combinations, obtaining 95.6\% accuracy
for 50 character sequences when choosing between 13
languages. \cite{acs1} used TF-IDF scores of words to distinguish
between language groups. Recently, the frequency of words has also been
used for \langid by \cite{clematide1}, \cite{gomezadorno1},
\cite{cagigos1}, and \cite{saharia1}.

\paragraph{The relative frequency of words}
\cite{poutsma1} and \cite{zhdanova1} were the first to use relative frequencies of words in \langid. As did \cite{prager1} for word frequencies, also
\cite{jauhiainen1} found that combining the use of character \ngrams
with the use of words provided the best results. His language identifier
obtained 99.8\% average recall for 50 character sequences for the 10
evaluated languages (choosing between the 13 languages known by the
language identifier) when using character \ngrams from 1 to 6 combined
with words. \cite{tiedemann1} calculated the relative frequency of
words over all the languages. \cite{artemenko2} calculated the IDF of
words, following the approach outlined in \eqnref{artemenko1}. \cite{xu1} calculated the Pointwise Mutual Information (``PMI'') for words and used it to group words to Chinese
dialects or dialect groups. Recently, the relative frequency of words
has also been used for \langid by \cite{jauhiainen7,jauhiainen6} and
\cite{jourlin1}

\paragraph{Short words}

\cite{grefenstette1} used the relative frequency of words with less
than six characters. Recently, \cite{panich1} also used short words, as did \cite{simaki1}.

\paragraph{Search engine queries}

\cite{alex1} used the relative frequency calculated from Google searches. Google was later also used by \cite{you1} and \cite{yang1}.

\paragraph{Word probability maps}

\cite{scherrer1} created probability maps for words for German dialect identification between six dialects. In a word probability map, each predetermined geographic point has a probability for each word form. Probabilities were derived using a linguistic atlas and automatically-induced dialect lexicons.


\paragraph{Morphological analyzers and spellchecking}

\cite{pienaar1} used commercial spelling checkers, which utilized
lexicons and morphological analyzers. The language identifier of
\cite{pienaar1} obtained 97.9\% accuracy when classifying one-line
texts between 11 official South African languages. \cite{elfardy1} used the
ALMORGEANA analyzer to check if the word had an analysis in modern
standard Arabic. They also used sound change rules to use possible
phonological variants with the analyzer. \cite{joshi1} used
spellchecking and morphological analyzers to detect English words from
Hindi--English mixed search queries. \cite{akosu3} used spelling
checkers to distinguish between 15 languages, extending the work of
\cite{pienaar1} with dynamic model selection in order to gain better
performance. \cite{shrestha1} used a similarity count to find if
mystery words were misspelled versions of words in a dictionary.

\paragraph{Word clusters}

\cite{pham1} used an ``LBG-VQ'' (Linde, Buzo \& Gray algorithm for Vector Quantization) approach to design a codebook for each language \citep{linde1}. The codebook contained a predetermined number of codevectors. Each codeword represented the word it was generated from as well as zero or more words close to it in the vector space.


\subsection{Word Combinations}

\paragraph{Sentence length}

\cite{elfardy2} used the number of words in a sentence with NB. \cite{lee3} and \cite{simaki1} used the sentence length calculated in both words and characters with several machine learning algorithms.

\paragraph{Statistics of words}

\cite{lee3} used the ratio to the total number of words of: once-occurring words, twice-occurring words, short words, long words, function words, adjectives and adverbs, personal pronouns, and question words. They also used the word-length distribution for words of 1--20 characters.

\paragraph{Word \ngrams}

\cite{marcadet1} used at least the preceding and proceeding words with
manual rules in word-level \langid for text-to-speech
synthesis. \cite{rosner1} used Markovian word \ngrams with a Hidden
Markov Model (``HMM'') tagger
(\secref{othermethods}). \tabref{WordNgramTable} lists more recent
articles where word \ngrams or similar constructs have been
used. ``PPM'' in the methods column refers to Prediction by Partial
Matching (\secref{smoothing}), and ``kNN'' to $k$ Nearest Neighbor classification (\secref{ensemble}).

\cite{singh1} used word trigrams simultaneously with character
4-grams. He concluded that word-based models can be used to augment the
results from character \ngrams when they are not providing reliable
identification results. \tabref{WordCharacterNgramTable} lists 
articles where both character and word \ngrams have been used
together. ``CBOW'' in the methods column refer to Continuous Bag of
Words neural network (\secref{neuralnetworks}), and ``MIRA'' to Margin Infused Relaxed Algorithm
(\secref{supportvectormachines}). \cite{sanchezperez1} evaluated
different combinations of word and character \ngrams with SVMs. The best
combination for language variety identification was using all the
features simultaneously. \cite{tellez1} used normal and gapped word
\ngrams and character \ngrams simultaneously.

\paragraph{Co-occurrences of words}
\label{sec:cooccurrencesofwords}

\cite{wan1} uses word embeddings consisting of Positive Pointwise Mutual Information (``PPMI'') counts to represent each word type. Then they use Truncated Singular Value Decomposition (``TSVD'') to reduce the dimension of the word vectors to 100. \cite{elgabou1}
used $k$-means clustering when building dialectal Arabic
corpora. \cite{kheng1} used features provided by Latent Semantic
Analysis (``LSA'') with SVMs and NB.

\cite{mikolov1} present two models, the CBOW model and the continuous skip-gram model. The CBOW model can be used to generate a word given it's context and the skip-gram model can generate the context given a word. The projection matrix, which is the weight matrix between the input layer and the hidden layer, can be divided into vectors, one vector for each word in the vocabulary. These word-vectors are also referred to as word embeddings. The embeddings can be used as features in other tasks after the neural network has been trained.
\cite{lin2}, \cite{chang1}, \cite{francosalvador2,francosalvador1}, \cite{jain2}, \cite{francosalvador3,francosalvador4}, and \cite{rangel1} used word embeddings generated by the word2vec skip-gram model \citep{mikolov1} as features in \langid. 
\cite{poulston1} used word2vec word embeddings and $k$-means clustering. \cite{akhtyamova1}, \cite{kodiyan1}, and \cite{samih3} also used word embeddings created with word2vec.
 
\cite{coltekin1} trained both character and word embeddings using FastText text classification method \citep{joulin1} on the Discriminating between Similar Languages (``DSL'') 2016 shared task, where it reached low accuracy when compared with the other methods.
\cite{xia5} used FastText to train word vectors including subword information. Then he used these word vectors together with some additional word features to train a CRF-model which was used for codeswitching detection.

\begin{table}[H]
\footnotesize
\begin{tabular}{p{5cm}lllllllp{3cm}}
\hline
\textbf{Article} & \textbf{1} & \textbf{2} & \textbf{3}	 & \textbf{4} & \textbf{5} & \textbf{6} & \textbf{7} & \textbf{Method}\\
\hline
\cite{bhattu1} & \bestone & \bestone & \bestone & \bestone & \bestone & & & LR (cf.\ \ssecref{maxent})\\
\cite{bobicev2} & & \bestone & & & & & & PPM-C (cf.\ \ssecref{smoothing})\\
\cite{ghosh1} & \bestone & \bestone & \bestone & \bestone & \bestone & \bestone & \bestone & CRF (cf.\ \ssecref{openissues:short})\\
\cite{huang1} & \bestone &\bestother&\bestother& & & & & Product (cf.\ \ssecref{product})\\
\cite{qiao1} &\bestOther&\bestOther& & & & & & SVM (cf.\ \ssecref{supportvectormachines})\\
\cite{raghavi1} & \bestone & \bestone & \bestone & \bestone & \bestone & \bestone & \bestone & SVM\\
\cite{shah1} & & \bestone & \bestone & \bestone & \bestone & & & SVM \\
\cite{adouane1,adouane3} & \bestone & \besttwo & \besttwo &\bestother&\bestother& & & SVM \\
\cite{adouane2} & \bestone & \bestone & \besttwo &\bestother& & & & NB (cf.\ \ssecref{product}), SVM, LR, kNN (cf.\ \ssecref{ensemble}), DT (cf.\ \ssecref{decisiontrees}) \\
\cite{barbaresi1} & & \bestone & & & & & & RF (cf.\ \ssecref{decisiontrees})\\
\cite{eldesouki1} & \bestone &\bestother&\bestother& & & & & SVM, LR, NN (cf.\ \ssecref{neuralnetworks}), NB\\
\cite{francopenya1} &\bestother& \bestone & & & & & & NB, SVM\\
\cite{hanani1} & \bestone & \bestone & \bestone & & & & & LSTM RNN (cf.\ \ssecref{neuralnetworks})\\
\cite{samih2,samih3} & \bestone & \bestone & \bestone & \bestone & \bestone & \bestone & \bestone & CRF\\
\cite{samih1} & \bestone & \bestone & \bestone & \bestone & \bestone & & & LSTM RNN - CRF\\
\cite{schulz1} & \bestone & \bestone & \bestone & \bestone & \bestone & & & CRF\\
\cite{sikdar1} & \bestone & \bestone & \bestone & \bestone & \bestone & & & CRF\\
\cite{xia5} & \bestone & \bestone & \bestone & & & & & CRF\\
\cite{zampieri10} & \bestone & \besttwo & & & & & & SVM\\
\cite{alrifai1} &\bestother&\bestother&\bestother& & & & & SVM\\
\cite{criscuolo1} & \bestone & \bestone & & & & & & NN (cf.\ \ssecref{neuralnetworks})\\
\cite{gamallo3} & \bestone & \bestone & \bestone & & & & & Perplexity (cf.\ \ssecref{perplexity})\\
\cite{hanani2} & \bestone & \bestone &\bestOther& & & & & SVM\\
\cite{kheng1} & \bestone & \bestone & \bestone & & & & & NB, SVM, RF\\
\cite{lee3} & & & & & & & & AdaBoost (cf.\ \ssecref{ensemble}), DT, SVM, NB, ...\\
\cite{mathur1} & \besttwo & \bestone &\bestother&\bestother&\bestother&\bestother&\bestother& NB, LR\\
\cite{mendoza1} & & \bestone & & & & & & Product\\
\cite{miura1} & \bestone & \bestone & & & & & & SVM\\
\cite{poulston1} & \bestone & \bestone & & & & & & LR\\
\cite{rangel1} & & \bestone & & & & & & SVM, NB, NN, ...\\
\cite{rijhwani1} & \bestone & \bestone & \bestone & & & & & HMM (cf.\ \ssecref{othermethods})\\
\cite{sanchezperez1} & \bestone & \bestone & & & & & & SVM\\
\cite{tellez1} & \bestone & \bestone & \bestone & & & & & SVM 
\\
\hline
\end{tabular}
\caption{References (2015--) where word \ngrams have been used as
  features. The numbered columns indicate the length of the \ngrams
  used. ``\bestone'' indicates the best and ``\besttwo'' the second best
  \ngram length, as evaluated in the article in question. ``\bestother''
  indicates that there was no clear order of effectiveness, or that the
  order was not presented in the article. The method column indicates
  the method used. The relevant section numbers are mentioned in
  parentheses.}
\label{tab:WordNgramTable}
\end{table}

\begin{table}[t!]
\footnotesize
\begin{tabular}{p{5cm}llllllllp{3,5cm}}
\hline
\textbf{Article} & \textbf{1} & \textbf{2} & \textbf{3}	 & \textbf{4} & \textbf{5} & \textbf{6} & \textbf{7} & \textbf{char}& \textbf{Method}\\
\hline
\cite{singh1,singh3} & & & \checkmark & & & & & 1-4 & similarity measures (cf.\ \ssecref{vectors})\\
\cite{das2,das1} & \checkmark & \checkmark & \checkmark & \checkmark & \checkmark & \checkmark & \checkmark & 1-7 & SVM (cf.\ \ssecref{supportvectormachines})\\
\cite{nguyen1} & \checkmark & \checkmark & \checkmark & & & & & 1-5 & LR (cf.\ \ssecref{maxent}), CRF (cf.\ \ssecref{openissues:short})\\
\cite{chittaranjan1} & \checkmark & \checkmark & \checkmark & & & & & 1-5 & CRF\\
\cite{darwish1} & \checkmark & \checkmark & \checkmark & & & & & 1-5 & RF (cf.\ \ssecref{decisiontrees})\\
\cite{goutte1} & \checkmark & \checkmark & & & & & & 2-6 & SVM, NB-like (cf.\ \ssecref{product})\\
\cite{gupta1} & \checkmark & \checkmark & \checkmark & \checkmark & \checkmark & & & 1-3 & RF, SVM, DT (cf.\ \ssecref{decisiontrees}), ...\\
\cite{king3} & \checkmark & \checkmark & & & & & & 1-5 & NB, LR, SVM\\
\cite{ullman1} & \checkmark & \checkmark & & & & & & 1-4 & NB-like\\
\cite{acs1} & \checkmark & \checkmark & & & & & & 1-4 & LR, SVM\\
\cite{chang1} & \checkmark & \checkmark & \checkmark & & & & & 2-3 & RNN (cf.\ \ssecref{neuralnetworks})\\
\cite{goutte2} & \checkmark & \checkmark & & & & & & 2-6 & SVM, NB-like\\
\cite{jain2} & \checkmark & \checkmark & \checkmark & \checkmark & \checkmark & \checkmark & \checkmark & 2-4 & CRF\\
\cite{malmasi1} & \checkmark & \checkmark & & & & & & 1-3 & SVM\\
\cite{malmasi2} & \checkmark & \checkmark & & & & & & 1-6 & SVM\\
\cite{malmasi3} & \checkmark & \checkmark & & & & & & 1-4 & SVM\\
\cite{castro1} & \checkmark & \checkmark & & & & & & 2-7 & Product (cf.\ \ssecref{product})\\
\cite{zirikly1} & \checkmark & \checkmark & \checkmark & & & & & 1-6 & LR\\
\cite{basile1} & \checkmark & \checkmark & & & & & & 3-6 & SVM\\
\cite{castro2} & \checkmark & \checkmark & & & & & & 2-7 & NB, LR, SVM, RF\\
\cite{ciobanu3} & \checkmark & \checkmark & & & & & & 1-6 & SVM\\
\cite{coltekin2} & \checkmark & \checkmark & \checkmark & & & & & 1-7+ & SVM\\
\cite{markov1} & \checkmark & \checkmark & \checkmark & & & & & 3-7 & SVM, NB\\
\cite{martinc1} & \checkmark & \checkmark & & & & & & 4 & SVM, LR, RF, ...\\
\cite{medvedeva1} & \checkmark & \checkmark & \checkmark & \checkmark & & & & 1-6 & SVM, CBOW (cf.\ \ssecref{neuralnetworks})\\
\cite{mendoza1} & \checkmark & \checkmark & & & & & & 1-6 & SVM\\
\cite{pla2} & \checkmark & \checkmark & \checkmark & \checkmark &  & & & 1-6 & SVM \\
\cite{williams1} & \checkmark & \checkmark & \checkmark & \checkmark & \checkmark & & & 1-5 & MIRA (cf.\ \ssecref{supportvectormachines})  \\
\hline
\end{tabular}
\caption{List of articles where word and character \ngrams have been used as features. The numbered columns indicate the length of the word \ngrams and char-column the length of character \ngrams used. The method column indicates the method used. The relevant section numbers are mentioned in parentheses.}
\label{tab:WordCharacterNgramTable}
\end{table}

\cite{barman2} extracted features from the hidden layer of a Recurrent Neural Network
(``RNN'') that had been trained to predict the next character in a string. They used the features with a SVM classifier.

\paragraph{Syntax and part-of-speech (``POS'') tags}

\cite{alex1} evaluated methods for detecting foreign language
inclusions and experimented with a Conditional Markov Model (``CMM'')
tagger, which had performed well on Named Entity Recognition
(``NER''). \cite{alex1} was able to produce the best results by
incorporating her own English inclusion classifier's decision as a
feature for the tagger, and not using the taggers POS
tags. \cite{romsdorfer2} used syntactic parsers together with
dictionaries and morpheme lexicons. \cite{lui4} used \ngrams composed
of POS tags and function words. \cite{piergallini2} used labels from a
NER system, cluster prefixes, and Brown clusters \citep{brown4}. 
\cite{adouane6} used POS tag \ngrams from one to three and
\cite{bestgen1} from one to five, and \cite{martinc1} used POS tag trigrams
with TF-IDF weighting. \cite{schulz1}, \cite{basile1}, \cite{lee3},
and \cite{simaki1} have also recently used POS
tags. \cite{francosalvador2} used POS tags with emotion-labeled
graphs in Spanish variety identification. In emotion-labeled graphs, each POS-tag was connected to one or more emotion nodes if a relationship between the original word and the emotion was found from the Spanish Emotion Lexicon. They also used POS-tags with IG-WP.
\cite{elfardy4} used the MADAMIRA tool for morphological analysis
disambiguation. The polySVOX text analysis module described by
\cite{romsdorfer2} uses two-level rules and morpheme lexicons on
sub-word level and separate definite clause grammars (DCGs) on word,
sentence, and paragraph levels. The language of sub-word units, words, sentences, and paragraphs in multilingual documents is identified at the same time as performing syntactic analysis for the document.
\cite{noh1} converted sentences into POS-tag patterns using a word-POS dictionary for Malay. The POS-tag patterns were then used by a neural network to indicate whether the sentences were written in Malay or not.
\cite{laboreiro1} used Jspell to detect differences in the grammar of
Portuguese variants. \cite{becavac1} used a syntactic grammar to
recognize verb-\emph{da}-verb constructions, which are characteristic of the Serbian language. The syntactic grammar was used together with several morphological grammars to distinguish between Croatian and Serbian.

\paragraph{Languages identified for surrounding words in word-level \langid}

\cite{marcadet1} used the weighted \langid scores of the words to the
left and right of the word to be classified. \cite{rosner1} used
language labels within an HMM. \cite{akhil1} used the language labels of
other words in the same sentence to determine the language of the
ambiguous word. The languages of the other words had been determined by
the positive Decision Rules (\secref{Decisionrule}), using dictionaries
of unique words when possible. \cite{das2,das1} used the language tags
of the previous three words with an SVM. \cite{mukherjee1} used language
labels of surrounding words with NB. \cite{king4} used the language
probabilities of the previous word to determining weights for
languages. \cite{king5} used unigram, bigram and trigram language label
transition probabilities. \cite{papalexakis1} used the language labels
for the two previous words as well as knowledge of whether
code-switching had already been detected or not. \cite{raj2} used the
language label of the previous word to determine the language of an
ambiguous word. \cite{sinha1} also used the language label of the
previous word. \cite{chanda2} used the language identifications of 2--4
surrounding words for post-identification correction in word-level
\langid. \cite{samih2} used language labels with a CRF. \cite{dongen1}
used language labels of the current and two previous words in
code-switching point prediction. Their predictive strength was lower 
than the count of code-switches, but better than the
length or position of the word. All of the features were used together
with NB, DT and SVM. \cite{guzman1} used language label bigrams with
an HMM. \cite{elfardy2} used the word-level language labels obtained with
the approach of \cite{elfardy3} on sentence-level dialect
identification.

\subsection{Feature Smoothing}
\label{sec:smoothing}

Feature smoothing is required in order to handle the cases where not all
features \(f_{i}\) in a test document have been attested in the training
corpora. Thus, it is used especially when the count of features is high,
or when the amount of training data is low. Smoothing is usually handled
as part of the method, and not pre-calculated into the language
models. Most of the smoothing methods evaluated by \cite{chen4} have
been used in \langid, and we follow the order of methods in that
article.

\paragraph{Additive smoothing (Laplace, Lidstone)}

In Laplace smoothing, an extra number of occurrences is added to every
possible feature in the language model. \cite{dunning1} used Laplace's
sample size correction (add-one smoothing) with the product of Markovian
probabilities. \cite{adams1} experimented with additive smoothing of
0.5, and noted that it was almost as good as Good-Turing
smoothing. \cite{chen4} calculate the values for each \ngram as:
\begin{equation}
v_{C_g}(f)=\frac{c(C_g,f)+\lambda}{l_{C_{g}^{n}}+|U(C_g^{n})|\lambda}
\label{eqn:KingLidstone}
\end{equation}
where \(v_{C_g}(f)\) is the probability estimate of \ngram \(f\) in the
model and \(c(C_g,f)\) its frequency in the training
corpus. \(l_{C_{g}^{n}}\) is the total number of \ngrams of length \(n\)
and \(|U(C_g^{n})|\) the number of distinct \ngrams in the training
corpus. \(\lambda\) is the Lidstone smoothing parameter. When using
Laplace smoothing, \(\lambda\) is equal to 1 and with Lidstone
smoothing, the \(\lambda\) is usually set to a value between 0 and 1.

The penalty values used by \cite{jauhiainen5} with the HeLI method
function as a form of additive smoothing. \cite{vatanen1} evaluated
additive, Katz, absolute discounting, and Kneser-Ney smoothing methods. Additive
smoothing produced the least accurate results of the four
methods. \cite{cann1} and \cite{francopenya1} evaluated NB with
several different Lidstone smoothing values. \cite{cianflone1} used
additive smoothing with character \ngrams as a baseline classifier,
which they were unable to beat with Convolutional Neural Networks
(``CNNs'').

\paragraph{Good-Turing Discounting}

\cite{adams1} used Good-Turing smoothing with the product of Markovian probabilities. \cite{chen4} define the Good-Turing smoothed count \(c_{GT}(C,f)\) as:
\begin{equation}
c_{GT}(C_g,f)=(c(C_g,f)+1)\frac{r_{c(C_g,f)+1}}{r_{c(C_g,f)}}
\label{eqn:GoodTuring}
\end{equation}
where \(r_{c(C_g,f)}\) is the number of features occurring exactly \(c(C_g,f)\) times in the corpus \(C_g\). Lately Good-Turing smoothing has been used by \cite{gamallo2} and \cite{giwa3}.

\paragraph{Jelinek-Mercer}

\cite{rehurek1} used Jelinek-Mercer smoothing correction over the
relative frequencies of words, calculated as follows:
\begin{equation}
\label{JelinekMercer}
v_{C_g}(f)=\lambda\frac {c(C,f)} {l_{C^{F}}} + (1-\lambda)\frac {c(C_g,f)} {l_{C_g^{F}}}
\end{equation}
where \(\lambda\) is a smoothing parameter, which is usually some small value like 0.1. \cite{mendizabal1} used character 1--8 grams with Jelinek-Mercer smoothing. Their language identifier using character 5-grams achieved 3rd place (out of 12) in the TweetLID shared task constrained track.

\paragraph{Katz}

\cite{ramisch1} and \cite{vatanen1} used the Katz back-off smoothing \citep{katz1} from the SRILM toolkit, with perplexity. Katz smoothing is an extension of Good-Turing discounting. The probability mass left over from the discounted \ngrams is then distributed over unseen \ngrams via a smoothing factor. In the smoothing evaluations by \cite{vatanen1}, Katz smoothing performed almost as well as absolute discounting, which produced the best results. \cite{giwa1} evaluated Witten-Bell, Katz, and absolute discounting smoothing methods. Witten-Bell got 87.7\%, Katz 87.5\%, and absolute discounting 87.4\% accuracy with character 4-grams.

\paragraph{Prediction by Partial Matching (PPM/Witten-Bell)}

\cite{teahan2} used the PPM-C algorithm for \langid. PPM-C is basically
a product of Markovian probabilities with an escape scheme. If an unseen
context is encountered for the character being processed, the escape
probability is used together with a lower-order model probability. In
PPM-C, the escape probability is the sum of the seen contexts in the
language model. PPM-C was lately used by \cite{adouane4}. The PPM-D+
algorithm was used by \cite{celikel1}. \cite{bergsma1} and
\cite{mcnamee2} used a PPM-A variant. \cite{yamaguchi1} also used
PPM. The language identifier of \cite{yamaguchi1} obtained 91.4\%
accuracy when classifying 100 character texts between 277
languages. \cite{jaech1} used Witten-Bell smoothing with perplexity.

\cite{herman1} used a Chunk-Based Language Model (``CBLM''), which is similar to PPM models.

\paragraph{Absolute discounting}

\cite{vatanen1} used several smoothing techniques with Markovian probabilities. Absolute discounting from the VariKN toolkit performed the best. \cite{vatanen1} define the smoothing as follows: a constant \(D\) is subtracted from the counts \(c(C_g,u_{i-n+1,...,i})\) of all observed \ngrams \(u_{i-n+1,...,i}\) and the held-out probability mass is distributed between the unseen \ngrams in relation to the probabilities of lower order \ngrams \(P_{g}(u_{i}|u_{i-n+2,...,i-1})\), as follows:
\begin{equation}
P_{C_g}(u_{i}|u_{i-n+1,...,i-1})=\frac{c(C_g,u_{i-n+1,...,i})-D}{c(C_g,u_{i-n+1,...,i-1})}+\lambda_{u_{i-n+1,...,i-1}}P_{C_g}(u_{i}|u_{i-n+2,...,i-1})
\label{eqn:absolute}
\end{equation}
where \(\lambda_{u_{i-n+1,...,i-1}}\) is a scaling factor that makes the conditional distribution sum to one. Absolute discounting with Markovian probabilities from the VariKN toolkit was later also used by \cite{rodrigues1}, \cite{maier1}, and \cite{jauhiainen6}.

\paragraph{Kneser-Ney smoothing}

The original Kneser-Ney smoothing is based on absolute discounting with an added back-off function to lower-order models \citep{vatanen1}. \cite{chen4} introduced a modified version of the Kneser-Ney smoothing using interpolation instead of back-off. 
\cite{chen1} used the Markovian probabilities with Witten-Bell and
modified Kneser-Ney smoothing. \cite{giwa3}, \cite{balazevic1}, and
\cite{rijhwani1} also recently used modified Kneser-Ney
discounting. \cite{barbaresi1} used both original and modified
Kneser-Ney smoothings. In the evaluations of \cite{vatanen1}, Kneser-Ney
smoothing fared better than additive, but somewhat worse than the Katz
and absolute discounting smoothing. Lately \cite{samih2} also used
Kneser-Ney smoothing.

\cite{castro1,castro2} evaluated several smoothing techniques with
character and word \ngrams: Laplace/Lidstone, Witten-Bell, Good-Turing,
and Kneser-Ney. In their evaluations, additive smoothing with 0.1
provided the best results. Good-Turing was not as good as additive
smoothing, but better than Witten-Bell and Kneser-Ney
smoothing. Witten-Bell proved to be clearly better than Kneser-Ney.

\section{Methods}
\label{sec:methods}

In recent years there has been a tendency towards attempting to combine
several different types of features into one classifier or classifier
ensemble. Many recent studies use readily available classifier
implementations and simply report how well they worked with the feature
set used in the context of their study. There are many methods presented
in this article that are still not available as out of the box
implementations, however. There are many studies which have not been
re-evaluated at all, going as far back as \cite{mustonen1}. Our hope is
that this article will inspire new studies and many previously unseen
ways of combining features and methods. In the following sections, the
reviewed articles are grouped by the methods used for \langid.

\subsection{Decision Rules}
\label{sec:Decisionrule}

\cite{henrich1} used a positive Decision Rules with unique characters
and character \ngrams, that is, if a unique character or character
\ngram was found, the language was identified. The positive Decision Rule (unique features) for the test document \(M\) and the training corpus \(C_{g}\) can be formulated as follows:
\begin{equation}
\label{eqn:posdecisionrule}
R_{\text{\emph{DR}}^{+}}(g,M) = \left\{ 
  \begin{array}{l l}
    1 & \quad \text{, if } \exists f \in U(M): c(C_{g},f)>0 \wedge c(C_{j},u)=0 \wedge g \neq j\\
    0 & \quad \text{, otherwise}
  \end{array} \right.
\end{equation}
where \(U(M)\) is the set of unique features in \(M\), \(C_{g}\) is the corpus for language \(g\), and \(C_{j}\) is a corpus of any other language \(j\). 
Positive decision rules can also be used with non-unique features
when the decisions are made in a certain order. For example,
\cite{dongen1} presents the pseudo code for her dictionary lookup tool,
where these kind of decisions are part of an if-then-else statement
block. Her (manual) rule-based dictionary lookup tool works better for
Dutch--English code-switching detection than the SVM, DT, or CRF methods
she experiments with. The positive Decision Rule has also been used
recently by \cite{abainia2}, \cite{chanda1,chanda2}, \cite{guellil1},
\cite{gupta2}, \cite{he1}, and \cite{adouane6}.

In the negative Decision Rule, if a character or character combination that was found in \(M\) does not exist in a particular language, that language is omitted from further identification. The negative Decision Rule can be expressed as:
\begin{equation}
\label{eqn:negdecisionrule}
R_{\text{\emph{DR}}^{-}}(g,M) = \left\{ 
  \begin{array}{l l}
  0 & \quad \text{, if } \exists f \in U(M): c(C_{g},f)=0\\
  1 & \quad \text{, otherwise}
      \end{array} \right.
\end{equation}
where \(C_{g}\) is the corpus for language \(g\). The negative Decision Rule was first used by \cite{giguet2} in \langid.

\cite{alshutayri1} evaluated the JRIP classifier from the Waikato Environment for Knowledge Analysis (``WEKA''). JRIP is an implementation of the propositional rule learner. It was found to be inferior to the SVM, NB and DT algorithms.

In isolation the desicion rules tend not to scale well to larger numbers of
languages (or very short test documents), and are thus mostly used in
combination with other \langid methods or as a Decision Tree.

\subsection{Decision Trees}
\label{sec:decisiontrees}

\cite{hakkinen1} were the earliest users of Decision Trees (``DT'') in \langid. They used DT based on characters and their context without any frequency information. In training the DT, each node is split into child nodes according to an information theoretic optimization criterion. For each node a feature is chosen, which maximizes the information gain at that node. The information gain is calculated for each feature and the feature with the highest gain is selected for the node. In the identification phase, the nodes are traversed until only one language is left (leaf node). Later, \cite{ceylan1}, \cite{eskander1}, and \cite{moodley1} have been especially successful in using DTs.

Random Forest (RF) is an ensemble classifier generating many DTs. It has been succesfully used in \langid by \cite{jhamtani1}, \cite{darwish1}, \cite{ranjan1}, and \cite{malmasi8,malmasi7}.

\subsection{Simple Scoring}
\label{sec:Simplescoring}

In simple scoring, each feature in the test document is checked against
the language model for each language, and languages which contain that
feature are given a point, as follows:
\begin{equation}
\label{eqn:simple1}
R_{\text{\emph{simple}}}(g,M) = \sum_{i=1}^{l_{M^F}} \left\{
  \begin{array}{ll}
    1 & \quad\text{, if } f_{i} \in dom(O(C_{g}))\\
    0 & \quad\text{, otherwise}
  \end{array} \right.
\end{equation}
where \(f_{i}\) is the \(i\)th feature found in the test document
\(M\). The language scoring the most points is the winner. Simple
scoring is still a good alternative when facing an easy problem such as
preliminary language group identification. It was recently used for this
purpose by \cite{francosalvador1} with a basic dictionary. They
achieved 99.8\% accuracy when identifying between 6 language
groups. \cite{kadri2} use a version of simple scoring as a distance
measure, assigning a penalty value to features not found in a model. In
this version, the language scoring the least amount of points is the
winner. Their language identifier obtained 100\% success rate with
character 4-grams when classifying relatively large documents (from 1 to
3 kilobytes), between 10 languages. Simple scoring was also used lately
by \cite{balazevic1}, \cite{akosu2}, and \cite{duvenhage1}.

\subsection{Sum or Average of Values}
\label{sec:sum}

The sum of values can be expressed as:
\begin{equation}
\label{eqn:sumvalues1}
R_{\text{\emph{sum}}}(g,M)= \sum_{i=1}^{l_{M^F}} v_{C_{g}}(f_{i})
\end{equation}
where \(f_{i}\) is the \(i\)th feature found in the test document \(M\), and \(v_{C_{g}}(f_{i})\) is the value for the feature in the language model of the language \(g\). The language with the highest score is the winner.

The simplest case of \eqnref{sumvalues1} is when the text to be identified contains only one feature. An example of this is \cite{shrestha1} who used the frequencies of short words as values in word-level identification. For longer words, he summed up the frequencies of different-sized \ngrams found in the word to be identified. \cite{giwa2} first calculated the language corresponding to each graphone. They then summed up the predicted languages, and the language scoring the highest was the winner. When a tie occurred, they used the product of the Markovian graphone \ngrams. Their method managed to outperform SVMs in their tests.

\cite{henrich1} used the average of all the relative frequencies of the
\ngrams in the text to be identified. \cite{vogel1} evaluated several
variations of the LIGA algorithm introduced by
\cite{tromp2}. \cite{moodley1} and \cite{jauhiainen6} also used LIGA
and logLIGA methods. The average or sum of relative frequencies was also
used recently by \cite{abainia2} and \cite{martadinata1}.

\cite{ng3} summed up LFDF values (see \secref{characters}), obtaining 99.75\% accuracy when
classifying document sized texts between four languages using Arabic
script.  \cite{vitale1} calculates the score of the language for the
test document \(M\) as the average of the probability estimates of the
features, as follows:
\begin{equation}
\label{eqn:vitale3}
R_{\text{\emph{avg}}}(g,M)= \sum_{i=1}^{l_{M^F}} \frac {v_{C_{g}}(f_{i})}{l_{M^F}}
\end{equation}
where \(l_{M^F}\) is the number of features in the test document
\(M\). \cite{brown2} summed weighted relative frequencies of character
\ngrams, and normalized the score by dividing by the length (in
characters) of the test document. Taking the average of the terms in the
sums does not change the order of the scored languages, but it gives
comparable results between different lengths of test documents.

\cite{vega1,vega2} summed up the feature weights and divided them by
the number of words in the test document in order to set a threshold to
detect unknown languages. Their language identifier obtained 89\%
precision and 94\% recall when classifying documents between five
languages. \cite{el-shishiny1} used a weighting method combining
alphabets, prefixes, suffixes and words. \cite{elfardy1} summed up
values from a word trigram ranking, basic dictionary and morphological
analyzer lookup. \cite{akhil1} summed up language labels of the
surrounding words to identify the language of the current
word. \cite{becavac1} summed up points awarded by the presence of
morphological and syntactic features. \cite{gamallo1} used inverse rank
positions as values. \cite{acs1} computed the sum of keywords weighted
with TF-IDF. \cite{fabraboluda1} summed up the TF-IDF derived
probabilities of words.

\subsection{Product of Values}
\label{sec:product}

The product of values can be expressed as follows:
\begin{equation}
\label{eqn:productvalues1}
R_{\text{\emph{prod}}}(g,M)= \prod_{i}^{l_{M^F}} v_{C_{g}}(f_{i})
\end{equation}
where \(f_{i}\) is the \(i\)th feature found in test document \(M\), and
\(v_{C_{g}}(f_{i})\) is the value for the feature in the language model
of language \(g\). The language with the highest score is the
winner. Some form of feature smoothing is usually required with the
product of values method to avoid multiplying by zero.

\paragraph{Product of relative frequencies}

\cite{church1} was the first to use the product of relative frequencies
and it has been widely used ever since; recent examples include
\cite{castro1,castro2}, \cite{hanani2}, and \cite{jauhiainen6}. Some
of the authors use a sum of log frequencies rather than a product of
frequencies to avoid underflow issues over large numbers of features,
but the two methods yield the same relative ordering, with the proviso
that the maximum of multiplying numbers between 0 and 1 becomes the
minimum of summing their negative logarithms, as can be inferred from:
\begin{equation}
\label{eqn:multitosum}
R_{\text{\emph{logsum}}}(g,M) = -\log (R_{\text{\emph{prod}}}(g,M)) = -\log \prod_{i=1}^{l_{M^F}} v_{C_{g}}(f_{i}) = \sum_{i=1}^{l_{M^F}} -\log (v_{C_{g}}(f_{i}))
\end{equation}

\paragraph{Naive Bayes (NB)}
\label{sec:naivebayes}

When (multinomial\footnote{To the best of our knowledge, the
  multivariate Bernoulli version of NB has never been used for
  \langid. See \cite{giwa3} for a possible explanation.}) NB is used in
\langid, each feature used has a probability to indicate each
language. The probabilities of all features found in the test document
are multiplied for each language, and the language with the highest
probability is selected, as in \eqnref{productvalues1}. Theoretically the
features are assumed to be independent of each other, but in practice
using features that are functionally dependent can improve
classification accuracy \citep{peng1}.

NB implementations have been widely used for \langid, usually with a
more varied set of features than simple character or word \ngrams of the
same type and length. The features are typically represented as feature
vectors given to a NB classifier. \cite{mukherjee1} trained a NB
classifier with language labels of surrounding words to help predict the
language of ambiguous words first identified using an SVM. The language
identifier used by \cite{tan3} obtained 99.97\% accuracy with 5-grams
of characters when classifying sentence-sized texts between six language
groups. \cite{goutte1} used a probabilistic model similar to
NB. \cite{bhattu1} used NB and naive Bayes EM, which uses the
Expectation--Maximization (``EM'') algorithm in a semi-supervised setting to
improve accuracy. \cite{ljubesic4} used Gaussian naive Bayes (``GNB'',
i.e.\ NB with Gaussian estimation over continuous variables) from
scikit-learn.

\paragraph{Bayesian Network Classifiers}
\label{sec:augmentedbayesian}

In contrast to NB, in Bayesian networks the features are not assumed to
be independent of each other. The network learns the dependencies
between features in a training phase. \cite{fabraboluda1} used a
Bayesian Net classifier in two-staged (group first) \langid
over the open track of the DSL
2015 shared task. \cite{rangel1} similarly evaluated Bayesian Nets, but
found them to perform worse than the other 11 algorithms they tested.


\paragraph{Product of Markovian probabilities}

\cite{house1} used the product of the Markovian probabilities of
character bigrams. The language identifier created by
\cite{brown2,brown3}, ``whatlang'', obtains 99.2\% classification
accuracy with smoothing for 65 character test strings, when
distinguishing between 1,100 languages. The product of Markovian
probabilities has recently also been used by \cite{samih2} and
\cite{mendoza1}.

\paragraph{HeLI}

\cite{jauhiainen5} use a word-based backoff method called HeLI. Here,
each language is represented by several different language models, only
one of which is used for each word found in the test document. The
language models for each language are: a word-level language model, and
one or more models based on character \ngrams of order
1--\(n_{max}\). When a word that is not included in the word-level model
is encountered in a test document, the method backs off to using
character \ngrams of the size \(n_{max}\). If there is not even a partial
coverage here, the method backs off to lower order \ngrams and continues
backing off until at least a partial coverage is obtained (potentially all the way to
character unigrams). The \langid system of \cite{jauhiainen5}
implementing the HeLI method attained shared first place in the
closed track of the DSL 2016 shared task \citep{malmasi9}, and was the
best method tested by \cite{jauhiainen6} for test documents longer than
30 characters.

\subsection{Similarity Measures}
\label{sec:vectors}

\paragraph{Out-of-place method}
\label{sec:outofplace}

The well-known method of \cite{cavnar1} uses overlapping character
\ngrams of varying sizes based on words. The language models are created
by tokenizing the training texts for each language \(g\) into words, and
then padding each word with spaces, one before and four after. Each
padded word is then divided into overlapping character \ngrams of sizes
1--5, and the counts of every unique \ngram are calculated over
the training corpus. The \ngrams are ordered by frequency and \(k\) of the
most frequent \ngrams, \(f_{1}, ..., f_{k}\), are used as the domain of
the language model \(O(C_{g})\) for the language \(g\). The rank of an
\ngram \(f\) in language \(g\) is determined by the \ngram frequency in
the training corpus \(C_{g}\) and denoted \(\rank_{C_{g}}(f)\).

During \langid, the test document  \(M\) is treated in a similar way and a corresponding model \(O(M)\) of the K most frequent \ngrams is created. Then a distance score is calculated between the model of the test document and each of the language models. The value \(v_{C_{g}}(f)\) is calculated as the difference in ranks between \(\rank_{C_{g}}(f)\) and \(\rank_{M}(f)\) of the \ngram \(f\) in the domain \(dom(O(M))\) of the model of the test document. If an \ngram is not found in a language model, a special penalty value \(p\) is added to the total score of the language for each missing \ngram. The penalty value should be higher than the maximum possible distance between ranks.

\begin{equation}
\label{eqn:Cavnar}
v_{C_{g}}(f) = \left\{
  \begin{array}{l l}
    |\rank_{M}(f)-\rank_{C_{g}}(f)| & \quad \text{, if } f \in dom(O(C_{g}))\\
    p & \quad \text{, if } f \notin dom(O(C_{g}))
  \end{array} \right.
\end{equation}
The score \(R_{\text{\emph{CT}}}(g)\) for each language \(g\) is the sum of values, as in \eqnref{sumvalues1}. The language with the lowest score \(R_{\text{\emph{CT}}}(g)\) is selected as the identified language. The method is equivalent to Spearman's measure of disarray \citep{diaconis1}. The out-of-place method has been widely used in \langid literature as a baseline. In the evaluations of \cite{jauhiainen6} for 285 languages, the out-of-place method achieved an F-score of 95\% for 35-character test documents. It was the fourth best of the seven evaluated methods for test document lengths over 20 characters.

\paragraph{Local Rank Distance (``LRD'')}

Local Rank Distance \citep{ionescu2} is a measure of difference between two strings. LRD is calculated by adding together the distances identical units (for example character \ngrams) are from each other between the two strings. The distance is only calculated within a local window of predetermined length. \cite{ionescu1} and \cite{ionescu3} used LRD with a Radial Basis Function (``RBF'') kernel (see \secref{RBF}). For learning they experimented with both Kernel Discriminant Analysis (``KDA'') and Kernel Ridge Regression (``KRR''). \cite{francosalvador3} also used KDA.

\paragraph{Levenshtein distance}

\cite{pavan1} calculated the Levenshtein distance between the language models and each word in the mystery text. The similary score for each language was the inverse of the sum of the Levenshtein distances.
Their language identifier obtained 97.7\% precision when
classifying texts from two to four words between five languages. Later
\cite{guellil1} used Levenshtein distance for Algerian dialect
identification and \cite{gupta2} for query word identification.

\paragraph{Probability difference}

\cite{botha1}, \cite{botha3}, \cite{botha2}, and \cite{botha4} calculated the difference between probabilities as in Equation~\ref{eq:botha1}.

\begin{equation}
\label{eq:botha1}
R_{\text{\emph{diff}}}(g,M)=\sum_{i} (v_{M}(f_{i})-v_{C_{g}}(f_{i}))
\end{equation}

\vspace{2mm}

\noindent where \(v_{M}(f_{i})\) is the probability for the feature \(f_{i}\) in the mystery text and \(v_{C_{g}}(f_{i})\) the corresponding probability in the language model of the language \(g\). The language with the lowest score \(R(g)\) is selected as the most likely language for the mystery text.
\cite{singh1,singh3} used the log probability difference
and the absolute log probability difference. The log probability
difference proved slightly better, obtaining a precision of 94.31\%
using both character and word \ngrams when classifying 100 character
texts between 53 language-encoding pairs.

\paragraph{Vectors}

Depending on the algorithm, it can be easier to view language models as
vectors of weights over the target features. In the following methods,
each language is represented by one or more feature vectors. Methods
where each feature type is represented by only one feature vector are also
sometimes referred to as centroid-based \citep{takci2} or nearest prototype methods.
Distance measures are generally applied to all features included in the
feature vectors.

\cite{kruengkrai2} calculated the squared Euclidean distance between feature vectors. The Squared Euclidean distance can be calculated as:
\begin{equation}
\label{eqn:sqeucdist1}
R_{\text{euc}^2}(g,M)=\sum_{i} (v_{M}(f_{i})-v_{C_{g}}(f_{i}))^2
\end{equation}
\cite{hamzah1} used the simQ similarity measure, which is closely
related to the Squared Euclidean distance.

\cite{stensby1} investigated the \langid of multilingual documents
using a Stochastic Learning Weak Estimator (``SLWE'') method. In SLWE, the
document is processed one word at a time and the language of each word
is identified using a feature vector representing the current word as
well as the words processed so far. This feature vector includes all
possible units from the language models -- in their case mixed-order character \ngrams from one to four. The vector is updated using the SLWE updating scheme to increase the probabilities of units found in the current word. The probabilities of units that have been found in previous words, but not in the current one, are on the other hand decreased.
After processing each word, the distance of the feature vector to the probability distribution of each language is calculated, and the best-matching language is chosen as the language of the current word.
Their language identifier obtained 96.0\% accuracy when classifying
sentences with ten words between three languages. They used the
Euclidean distance as the distance measure as follows:
\begin{equation}
\label{eqn:eucdist1}
R_{\text{euc}}(g,M)=\sqrt{R_{\text{euc}^2}(g,M)}
\end{equation}

\cite{tomovic1} compared the use of Euclidean distance with their
own similarity functions.  \cite{prager1} calculated the cosine angle
between the feature vector of the test document and the feature vectors
acting as language models. This is also called the cosine similarity and
is calculated as follows:
\begin{equation}
\label{eqn:cosangle1}
R_{\text{cos}}(g,M)=\frac {\sum_{i} v_{M}(f_{i})v_{C_{g}}(f_{i})} {\sqrt{\sum_{i} v_{M}(f_{i})^2}\sqrt{\sum_{i} v_{C_{g}}(f_{i})^2}}
\end{equation}
The method of \cite{prager1} was evaluated by \cite{lui3} in the
context of \langid over multilingual documents. The cosine similarity
was used recently by \cite{schaetti1}. One common trick with cosine
similarity is to pre-normalise the feature vectors to unit length (e.g.\
\cite{brown1}), in
which case the calculation takes the form of the simple dot
product:\begin{equation}
\label{eqn:dotprod1}
R_{\text{dotprod}}(g,M)={\sum_{i} v_{M}(f_{i})v_{C_{g}}(f_{i})}
\end{equation}

\cite{jauhiainen1} used chi-squared distance, calculated as follows:
\begin{equation}
\label{eqn:chi-squared}
R_{\text{\emph{chi-square}}}(g,M)=\sum_{i} \frac{(v_{C_{g}}(f_{i})-v_{M}(f_{i}))^{2}}{v_{M}(f_{i})}
\end{equation}

\cite{abainia2} compared Manhattan, Bhattacharyya, chi-squared,
Canberra, Bray Curtis, histogram intersection, correlation distances,
and out-of-place distances, and found the out-of-place method to be the most accurate.

\paragraph{Entropy}

\cite{singh1,singh3} used cross-entropy and symmetric
cross-entropy. Cross-entropy is calculated as follows,
where \(v_{M}(f_{i})\) and \(v_{C_{g}}(f_{i})\) are the probabilities of
the feature \(f_{i}\) in the the test document \(M\) and the corpus
\(C_{g}\):
\begin{equation}
\label{eqn:crossentropy}
R_{\text{\emph{cross-entropy}}}(g,M)=\sum_{i} v_{M}(f_{i}) \log v_{C_{g}}(f_{i})
\end{equation}
Symmetric cross-entropy is calculated as:
\begin{equation}
\label{eqn:symcrossentropy}
R_{\text{\emph{sym-cross-entropy}}}(g,M)=\sum_{i} v_{M}(f_{i}) \log v_{C_{g}}(f_{i})+v_{C_{g}}(f_{i}) \log v_{M}(f_{i})
\end{equation}
For cross-entropy, distribution $M$ must be smoothed, and for symmetric
cross-entropy, both probability distributions must be smoothed.
Cross-entropy was used recently by \cite{hanani2}. \cite{yamaguchi1} used a cross-entropy estimating method they call the Mean of Matching Statistics (``MMS''). In MMS every possible suffix of the mystery text \(u_i,...,u_{l_M}\) is compared to the language model of each language and the average of the lengths of the longest possible units in the language model matching the beginning of each suffix is calculated.

\cite{sibun1} and \cite{baldwin2} calculated the relative entropy between the language models and the test document, as follows:
\begin{equation}
\label{eqn:kullback1}
R_{\text{\emph{rel-entropy}}}(g,M)=\sum_{i} v_{M}(f_{i}) \log \frac {v_{M}(f_{i})}{v_{C_{g}}(f_{i})}
\end{equation}
This method is also commonly referred to as Kullback-Leibler (``KL'')
distance or skew divergence. \cite{jauhiainen1} compared relative entropy with the product
of the relative frequencies for different-sized character \ngrams, and
found that relative entropy was only competitive when used with
character bigrams. The product of relative frequencies gained clearly higher recall with higher-order \ngrams when compared with relative entropy.



\cite{singh1,singh3} also used the RE and MRE measures, which are based
on relative entropy. The RE measure is calculated as follows:
\begin{equation}
\label{eqn:re-measure}
R_{\text{\emph{RE}}}(g,M)=\sum_{i} v_{M}(f_{i}) \frac {\log v_{M}(f_{i})}{\log v_{C_{g}}(f_{i})}
\end{equation}
MRE is the symmetric version of the same measure. In the tests performed by \cite{singh1,singh3}, the RE measure with character \ngrams outperformed other tested methods obtaining 98.51\% precision when classifying 100 character texts between 53 language-encoding pairs.

\paragraph{Logistic Regression (LR)}
\label{sec:maxent}

\cite{chen1} used a logistic regression (``LR'') model (also commonly
referred to as ``maximum entropy'' within NLP), smoothed with a Gaussian prior. \cite{porta1} defined LR for character-based features as follows:
\begin{equation}
\label{eqn:maxent}
R_{\text{\emph{LR}}}(g,M)=\frac{1}{Z} \exp \sum_{j}^{l_{M^T}} \sum_{i}^{l_{C^F}}  v_{C_{g}}(f_{i}) \text{, if } \exists f_{i} \in U(M_j^T)
\end{equation}
where \(Z\) is a normalization factor and \(l_{M^T}\) is the word count in
the word-tokenized test document. \cite{acs1} used an LR classifier and
found it to be considerably faster than an SVM, with comparable
results. Their LR classifier ranked 6 out of 9 on the closed submission
track of the DSL 2015 shared task. \cite{lu1} used Adaptive Logistic Regression, which automatically optimizes parameters.
In recent years LR has been widely used for \langid.

\paragraph{Perplexity}
\label{sec:perplexity}

\cite{ramisch1} was the first to use perplexity for \langid, in the
manner of a language model. He calculated the perplexity for the test document \(M\) as follows:

\begin{equation}
\label{eqn:ramisch}
H_{g}(M) = \frac{1}{l_{M^n}} \sum_{i}^{l_{M^n}} \log_{2} v_{C_{g}}(f_{i})
\end{equation}

\begin{equation}
R_{\text{\emph{perplexity}}}(g,M)=2^{H_{g}(M)}
\label{eqn:perplexity}
\end{equation}
where \(v_{C_{g}}(f_{i})\) were the Katz smoothed relative frequencies of word \emph{n}-grams \(f_{i}\) of the length \(n\).
\cite{rodrigues1} and \cite{jauhiainen6} evaluated the best performing method used by \cite{vatanen1}. Character \emph{n}-gram based perplexity was the best method for extremely short texts in the evaluations of \cite{jauhiainen6}, but for longer sequences the methods of \cite{brown1} and \cite{jauhiainen1} proved to be better. Lately, \cite{gamallo3} also used perplexity. 

\paragraph{Other similarity measures}

\cite{rau1} used Yule's characteristic K and the Kolmogorov-Smirnov
goodness of fit test to categorize languages. Kolmogorov-Smirnov proved
to be the better of the two, obtaining 89\% recall for 53 characters
(one punch card) of text when choosing between two languages. In the
goodness of fit test, the ranks of features in the models of the
languages and the test document are compared. \cite{martins1}
experimented with Jiang and Conrath's (JC) distance \citep{jiang1} and Lin's similarity
measure \citep{lin1}, 
as well as the out-of-place method. They conclude that Lin's similarity
measure was consistently the most accurate of the three. JC-distance
measure was later evaluated by \cite{singh1,singh3}, and was
outperformed by the RE measure. \cite{bali2} and \cite{bali1}
calculated special ratios from the number of trigrams in the language
models when compared with the text to be
identified. \cite{dasilva3,dasilva1,dasilva2} used the quadratic
discrimination score to create the feature vectors representing the
languages and the test document. They then calculated the Mahalanobis
distance between the languages and the test document. Their language
identifier obtained 98.9\% precision when classifying texts of four
``screen lines'' between 19 languages. 
\cite{nguyen3} used odds ratio to identify the language of parts of
words when identifying between two languages. Odds ratio for language \(g\) when compared with language \(h\) for morph \(f_i\) is calculated as in Equation~\ref{eq:odds}.

\begin{equation}
\label{eq:odds}
R_{\text{\emph{odds}}}(g,f_i)= \log \frac{v_{C_{h}}(f_{i}) (1-v_{C_{g}}(f_{i}))}{(1-v_{C_{h}}(f_{i}))v_{C_{g}}(f_{i})}
\end{equation}


\subsection{Discriminant Functions}
\label{sec:discriminantfunctions}

The differences between languages can be stored in discriminant
functions. The functions are then used to map the test document into an
$n$-dimensional space. The distance of the test document to the
languages known by the language identifier is calculated, and the
nearest language is selected (in the manner of a nearest prototype classifier).

\cite{murthy1} used multiple linear regression to calculate
discriminant functions for two-way \langid for Indian
languages. \cite{bhargava2} compared linear regression, NB, and LR. The
precision for the three methods was very similar, with linear regression
coming second in terms of precision after LR.

Multiple discriminant analysis was used for \langid by
\cite{mustonen1}. He used two functions, the first separated Finnish
from English and Swedish, and the second separated English and Swedish
from each other. He used Mahalanobis' \(D^2\) as a distance
measure. \cite{vinosh1} used Multivariate Analysis (``MVA'') with Principal Component Analysis (``PCA'') for dimensionality reduction and \langid. \cite{takci5} compared discriminant analysis with SVM and NN
using characters as features, and concluded that the SVM was the best method.


\cite{king1} experimented with the Winnow 2 algorithm \citep{littlestone1}, but the method
was outperformed by other methods they tested. 

\subsection{Support Vector Machines (``SVMs'')}
\label{sec:supportvectormachines}

With support vector machines (``SVMs''), a binary classifier is learned by
learning a separating hyperplane between the two classes of instances
which maximizes the margin between them. The simplest way to extend the
basic SVM model into a multiclass classifier is via a suite of
one-vs-rest classifiers, where the classifier with the highest score determines the language of the test
document. One feature of SVMs that has made them particularly popular is
their compatibility with kernels, whereby the separating hyperplane can
be calculated via a non-linear projection of the original instance
space. In the following paragraphs, we list the different kernels that
have been used with SVMs for \langid.

\paragraph{Linear kernel SVMs}

For \langid with SVMs, the predominant approach has been a simple
linear kernel SVM model. The linear kernel model has a weight vector \(v_{C_{g}}(f)\) and the classification of a feature vector \(v_{M}(f)\), representing the test document \(M\), is calculated as follows:
\begin{equation}
\label{eqn:linearSVM}
R_{\text{svm-lin}}(g,M)=(\sum_{i} v_{M}(f_{i})v_{C_{g}}(f_{i}))+b
\end{equation}
where \(b\) is a scalar bias term. If \(R_{\text{svm-lin}}\) is equal to or greater than zero,
\(M\) is categorized as \(g\).

The first to use a linear kernel SVM were \cite{kim1}, and generally
speaking, linear-kernel SVMs have been widely used for \langid, with
great success across a range of shared tasks.

\paragraph{Polynomial kernel SVMs}

\cite{bar1} were the first to apply polynomial kernel SVMs to \langid. With a polynomial kernel \(R_{\text{svm-pol}}\) can be calculated as:
\begin{equation}
\label{eqn:polynomialSVM}
R_{\text{svm-pol}}(g,M)=((\sum_{i} v_{M}(f_{i})v_{C_{g}}(f_{i}))+b)^d
\end{equation}
where \(d\) is the polynomial degree, and a hyperparameter of the model.

\paragraph{Radial Basis Function (RBF) kernel SVMs}
\label{sec:RBF}

Another popular kernel is the RBF function, also known as a Gaussian or
squared exponential kernel. With an RBF kernel \(R_{\text{svm-rbf}}\) is calculated as:
\begin{equation}
\label{eqn:RBFSVM}
R_{\text{svm-rbf}}(g,M)=\exp(-\frac{(\sum_{i} |v_{M}(f_{i})-v_{C_{g}}(f_{i})|)^2}{2\sigma^2})
\end{equation}
where \(\sigma\) is a hyperparameter. \cite{botha1} were the first to use an RBF kernel SVM for \langid. 

\paragraph{Sigmoid kernel SVMs}

With sigmoid kernel SVMs, also known as hyperbolic tangent SVMs, \(R_{\text{svm-sig}}\) can be calculated as:
\begin{equation}
\label{eqn:sigmoidSVM}
R_{\text{svm-sig}}(g,M)=\tanh((\sum_{i} v_{M}(f_{i})v_{C_{g}}(f_{i}))+b)
\end{equation}
\cite{bhargava1} were the first to use a sigmoid kernel SVM for
\langid, followed by \cite{majlis2}, who found the SVM to perform
better than NB, Classification And Regression Tree (``CART''), or the sum of relative frequencies.

\paragraph{Other kernels}

Other kernels that have been used with SVMs for \langid include
exponential kernels \citep{alrifai1} and rational kernels
\cite{porta2}. \cite{kruengkrai2} were the first to use SVMs for
\langid, in the form of string kernels using Ukkonen's
algorithm. They used same string kernels with Euclidean distance, which did not perform as well as SVM.
 \cite{castro2} compared SVMs
with linear and on-line passive--aggressive kernels for \langid, and
found passive--aggressive kernels to perform better, but both SVMs to be
inferior to NB and Log-Likelihood Ratio (sum of
log-probabilities). \cite{kim1} experimented with the Sequential
Minimal Optimization (``SMO'') algorithm, but found a simple linear kernel
SVM to perform better. \cite{alshutayri1} achieved the best results
using the SMO algorithm, whereas \cite{lamabam1} found CRFs to work
better than SMO. \cite{alrifai1} found that SMO was better than linear,
exponential and polynomial kernel SVMs for Arabic tweet gender and
dialect prediction.

\tabref{MultipleKernelSVMarticlesTable} lists articles where SVMs with
different kernels have been compared. \cite{goutte3} evaluated three
different SVM approaches using datasets from different DSL shared
tasks. SVM-based approaches were the top performing systems in the 2014
and 2015 shared tasks.

\begin{table}[t!]
\footnotesize
\begin{tabular}{lcccccccc}
\hline
\textbf{Reference} & linear & string & RBF & sigmoid & \(d^n\) & exp. & pas.\ aggr.\\
\hline
\cite{bhargava1} & \bestone &\bestother& \besttwo & \besttwo & \bestother\\
\cite{takci2} &\bestOther& &\bestOther& \bestother\\
\cite{giwa1,giwa3} & \besttwo & & \bestone\\
\cite{eldesouki1} & \bestone &\bestother\\
\cite{hanani1} & \bestone & &\bestother& & \bestother\\
\cite{xu1} & \bestone & &\bestother&\bestother&\bestother\\
\cite{alrifai1} & \besttwo & & & & \bestone & \bestother\\
\cite{castro2} & \besttwo & & & & & & \bestone\\
\cite{francosalvador3} &\bestOther& &\bestOther& &\bestOther \\
\hline
\end{tabular}
\caption{References where SVMs have been tested with different kernels. The columns indicate the kernels used. ``\(d^n\)'' stands for polynomial kernel.}
\label{tab:MultipleKernelSVMarticlesTable}
\end{table}

\paragraph{Margin Infused Relaxed Algorithm (``MIRA'')}

\cite{williams1} used SVMs with the Margin Infused Relaxed Algorithm,
which is an incremental version of SVM training. In their evaluation, this
method achieved better results than off-the-shelf \ldpy.

\subsection{Neural Networks (``NN'')}
\label{sec:neuralnetworks}

\cite{batchelder1} was the first to use Neural Networks (``NN'') for
\langid, in the form of a simple BackPropagation Neural Network
(``BPNN'') \citep{hechtnielsen1} with a single layer of hidden units,
which is also called a multi-layer perceptron (``MLP'') model. She used
words as the input features for the neural network.
\cite{tian1} and \cite{tian2} succesfully applied MLP to \langid.

\cite{jalam1,jalam2} and \cite{jalam3} used radial basis function
(RBF) networks for \langid. \cite{selamat3} were the first to use
adaptive resonance learning (``ART'') neural networks for
\langid. \cite{abainia2} used Neural Text Categorizer (``NTC'':
\cite{jo1}) as a baseline. NTC is an MLP-like NN using string vectors
instead of number vectors.

\cite{macnamara1} were the first to use a RNN for \langid. They
concluded that RNNs are less accurate than the simple sum of logarithms
of counts of character bi- or trigrams, possibly due to the relatively
modestly-sized dataset they experimented with. \cite{babu1} compared
NNs with the out-of-place method (see sec. \ref{sec:outofplace}). Their
results show that the latter, used with bigrams and trigrams of
characters, obtains clearly higher identification accuracy when dealing
with test documents shorter than 400 characters.

RNNs were more successfully used later by \cite{chang1} who also
incorporated character \emph{n}-gram features in to the network
architecture. \cite{cazamias1} were the first to use a Long Short-Term
Memory (``LSTM'') for \langid \citep{hochreiter1}, and \cite{bjerva1}
was the first to use Gated Recurrent Unit networks (``GRUs''), both of
which are RNN variants. \cite{bjerva1} used byte-level representations
of sentences as input for the networks.
Recently, \cite{hanani1} and \cite{samih1} also used LSTMs. Later,
GRUs were successfully used for \langid by \cite{jurgens1} and
\cite{kocmi1}.
In addition to GRUs, \cite{bjerva1} also
experimented with deep residual networks (``ResNets'') at DSL 2016.

During 2016 and 2017, there was a spike in the use of convolutional
neural networks (CNNs) for \langid, most successfully by
\cite{jaech1} and \cite{jaech2}. Recently, \cite{Li+:2018a} combined a CNN with
adversarial learning to better generalize to unseen domains, surpassing
the results of \cite{lui2} based on the same training regime as \ldpy. 

\cite{medvedeva1} used CBOW NN, achieving better results over the
development set of DSL 2017 than RNN-based neural networks.
\cite{francosalvador4} used deep averaging networks (DANs) based on word
embeddings in language variety identification.



\subsection{Other Methods}
\label{sec:othermethods}

\cite{simaki1} used the decision table majority classifier algorithm
from the WEKA toolkit in English variety detection. The bagging
algorithm using DTs was the best method they tested (73.86\%
accuracy), followed closely by the decision table with 73.07\% accuracy.

\cite{ueda1} were the first to apply hidden Markov models (HMM) to
\langid. More recently HMMs have been used by \cite{adouane6},
\cite{guzman1}, and \cite{rijhwani1}. \cite{binas1} generated
aggregate Markov models, which resulted in the best results when distinguishing between six languages, obtaining 74\% accuracy with text length of ten characters. 
\cite{king5} used an extended Markov Model (``eMM''), which is essentially
a standard HMM with modified emission probabilities. Their eMM used
manually optimized weights to combine four \ngram scores (products of
relative frequencies) into one \ngram score. 
\cite{xia3} used Markov logic networks \citep{richardson1} to predict the language used in
interlinear glossed text examples contained in linguistic papers. 

\cite{hayta1} evaluated the use of unsupervised Fuzzy C Means algorithm
(``FCM'') in language identification. The unsupervised algorithm was
used on the training data to create document clusters. Each cluster was
tagged with the language having the most documents in the cluster. Then
in the identification phase, the mystery text was mapped to the closest
cluster and identified with its language. A supervised centroid
classifier based on cosine similarity obtained clearly better results in
their experiments (93\% vs. 77\% accuracy).

\cite{barbaresi1} and \cite{martinc1} evaluated the extreme gradient
boosting (``XGBoost'') method \citep{chen3}.
\cite{barbaresi1} found that gradient boosting gave better results than
RFs, while conversely, \cite{martinc1} found that LR gave better
results than gradient boosting.

\cite{benedetto1} used compression methods for \langid, whereby a
single test document is added to the training text of each language in
turn, and the language with the smallest difference (after compression)
between the sizes of the original training text file and the combined
training and test document files is selected as the prediction. This has
obvious disadvantages in terms of real-time computational cost for
prediction, but is closely related to language modeling approaches to
\langid (with the obvious difference that the language model doesn't
need to be retrained multiply for each test document). In terms of
compression methods, \cite{hategan1} experimented with Maximal Tree
Machines (``MTMs''), and \cite{bush1} used LZW-based compression.


Very popular in text categorization and topic modeling, \cite{tratz2},
\cite{tratz1}, and \cite{voss2} used Latent Dirichlet Allocation
(``LDA'': \cite{Blei+:2003}) based features in classifying tweets
between Arabic dialects, English, and French. Each tweet was assigned
with an LDA topic, which was used as one of the features of an LR
classifier.



\cite{poulston1} used a Gaussian Process classifier with an RBF kernel
in an ensemble with an LR classifier. Their ensemble achieved only ninth
place in the ``PAN'' (Plagiarism Analysis, Authorship Identification, and
Near-Duplicate Detection workshop) Author Profiling language variety
shared task \citep{rangel2} and did not reach the results of the
baseline for the task.

\cite{linares1,linares2} used a Passive Aggressive classifier, which
proved to be almost as good as the SVMs in their evaluations between
five different machine learning algorithms from the same package.

\subsection{Ensemble Methods}
\label{sec:ensemble}

Ensemble methods are meta-classification methods capable of combining
several base classifiers into a combined model via a
``meta-classifier'' over the outputs of the base classifiers, either
explicitly trained or some heuristic. It is a simple and effective
approach that is used widely in machine learning to boost results beyond
those of the individual base classifiers, and particularly effective
when applied to large numbers of individually uncorrelated base
classifiers.

\paragraph{Majority and Plurality Voting}

\cite{rau1} used simple majority voting to combine classifiers using
different features and methods. In majority voting, the language of the
test document is identified if a majority (\(>\frac{1}{2}\)) of the
classifiers in the ensemble vote for the same language. In plurality
voting, the language with most votes is chosen as in the simple scoring
method (\eqnref{simple1}). Some authors also refer to plurality voting
as majority voting.

\cite{carter2} used majority voting in tweet \langid. \cite{giwa2}
used majority voting with JSM classifiers. \cite{goutte1} and
\cite{malmasi1} used majority voting between SVM classifiers trained
with different features. \cite{gupta1} used majority voting to combine
four classifiers: RF, random tree, SVM, and DT. \cite{doval1} and
\cite{lui5} used majority voting between three off-the-shelf language
identifiers. \cite{leidig2} used majority voting between
perplexity-based and other classifiers. \cite{zamora1} used majority
voting between three sum of relative frequencies-based classifiers where
values were weighted with different weighting
schemes. \cite{malmasi2,malmasi5}, \cite{malmasi4,malmasi8,malmasi7},
and \cite{mendoza1} used plurality voting with SVMs.

\cite{gamallo3} used voting between several perplexity-based
classifiers with different features at the 2017 DSL shared task. A
voting ensemble gave better results on the closed track than a singular
word-based perplexity classifier (0.9025 weighted F1-score over 0.9013),
but worse results on the open track (0.9016 with ensemble and 0.9065
without).

\paragraph{Highest Probability Ensemble}

In a highest probability ensemble, the winner is simply the language
which is given the highest probability by any of the individual
classifiers in the ensemble. \cite{you1} used Gaussian Mixture Models
(``GMM'') to give probabilities to the outputs of classifiers using
different features. \cite{doval1} used higher confidence between two
off-the-shelf language identifiers. \cite{goutte1} used GMM to
transform SVM prediction scores into
probabilities. \cite{malmasi2,malmasi5} used highest confidence over a
range of base SVMs. \cite{malmasi5} used an ensemble composed of
low-dimension hash-based classifiers. According to their experiments,
hashing provided up to 86\% dimensionality reduction without negatively
affecting performance. Their probability-based ensemble obtained 89.2\%
accuracy, while the voting ensemble got 88.7\%. \cite{balazevic1}
combined an SVM and a LR classifier.

\paragraph{Mean Probability Rule}

A mean probability ensemble can be used to combine classifiers that
produce probabilities (or other mutually comparable values) for
languages. The average of values for each language over the classifier
results is used to determine the winner and the results are equal to the
sum of values method (\eqnref{sumvalues1}). \cite{malmasi2} evaluated
several ensemble methods and found that the mean probability ensemble
attained better results than plurality voting, median probability,
product, highest confidence, or Borda count ensembles.

\paragraph{Median Probability Rule}

In a median probability ensemble, the medians over the probabilities given
by the individual classifiers are calculated for each
language. \cite{malmasi2} and \cite{malmasi4} used a median probability
rule ensemble over SVM classifiers. Consistent with the results of
\cite{malmasi2}, \cite{malmasi4} found that a mean ensemble was better
than a median ensemble, attaining 68\% accuracy vs.\ 67\% for the
median ensemble.

\paragraph{Product Rule}

A product rule ensemble takes the probabilities for the base classifiers
and calculates their product (or sum of the log probabilities), with the
effect of penalising any language where there is a particularly low
probability from any of the base classifiers. \cite{giwa2} used log
probability voting with JSM classifiers. \cite{giwa2} observed a small
increase in average accuracy using the product ensemble over a majority
voting ensemble.

\paragraph{$k$-best Ensemble ($k$-best)}

In a $k$-best ensemble, several models are created for each language \(g\) by
partitioning the corpus \(C_g\) into separate samples. The score
\(R(C_{g_i},M)\) is calculated for each model. For each language,
plurality voting is then applied to the $k$ models with the best scores to predict
the language of the test document \(M\).
\cite{jalam2} evaluated $k$-best with \(k=1\) based on several
similarity measures. \cite{kerwin1} compared \(k=10\) and \(k=50\) and
concluded that there was no major difference in accuracy when
distinguishing between six languages (100 character test
set). \cite{baykan1} experimented with $k$-best classifiers, but they
gave clearly worse results than the other classifiers they
evaluated. \cite{barman2} used $k$-best in two phases, first selecting
\(k_1=800\) closest neighbors with simple similarity, and then using
\(k_2=16\) with a more advanced similarity ranking.

\paragraph{Bootstrap Aggregating (Bagging)}

In bagging, independent samples of the training data are generated by
random sampling with replacement, individual classifiers are trained
over each such training data sample, and the final classification is
determined by plurality voting. \cite{martinc1} evaluated the use of
bagging with an LR classifier in PAN 2017 language variety
identification shared task, however, bagging did not improve the
accuracy in the 10-fold cross-validation experiments on the training
set.
\cite{sierra1} used bagging with word convolutional neural networks
(``W-CNN''). \cite{simaki1} used bagging with DTs in English national
variety detection and found DT-based bagging to be the best evaluated
method when all 60 different features (a wide selection of formal, POS,
lexicon-based, and data-based features) were used, attaining 73.86\%
accuracy. \cite{simaki1} continued the experiments using the ReliefF
feature selection algorithm from the WEKA toolkit to select the most
efficient features, and achieved 77.32\% accuracy over the reduced
feature set using a NB classifier.

\paragraph{Rotation Forest}

\cite{rangel1} evaluated the Rotation Forest meta classifier for
DTs. The method randomly splits the used features into a pre-determined
number of subsets and then uses PCA for each subset. It obtained 66.6\%
accuracy, attaining fifth place among the twelve methods evaluated.

\paragraph{Adaptive Boosting (AdaBoost)}

The AdaBoost algorithm \citep{freund2} examines the performance of the
base classifiers on the evaluation set and iteratively boosts the
significance of misclassified training instances, with a restart
mechanism to avoid local minima.
AdaBoost was the best of the five machine learning techniques evaluated
by \cite{lee3}, faring better than C4.5, NB, RF, and linear
SVM. \cite{rangel1} used the LogitBoost variation of AdaBoost. It
obtained 67.0\% accuracy, attaining third place among the twelve methods
evaluated.

\paragraph{Stacked generalization (Stacking)}

In stacking, a higher level classifier is explicitly trained on the output of
several base classifiers. \cite{you1} used AdaBoost.ECC and CART to
combine classifiers using different features. More recently,
\cite{mathur1} used LR to combine the results of five RNNs. As an
ensemble they produced better results than NB and LR, which were better
than the individual RNNs. Also in 2017, \cite{malmasi8,malmasi7} used
RF to combine several linear SVMs with different features. The system
used by \cite{malmasi7} ranked first in the German dialect
identification shared task, and the system by \cite{malmasi8} came
second (71.65\% accuracy) in the Arabic dialect identification shared
task.

\section{Empirical Evaluation}
\label{sec:evaluation}

In the previous two sections, we have alluded to issues of evaluation in \langid 
research to date. In this section, we examine the literature more closely,
providing a broad overview of the evaluation metrics that have been used, as well as the 
experimental settings in which \langid research has been evaluated.

\subsection{Standardized Evaluation for \langid}
\label{sec:standardevaluation}

The most common approach is to treat the task as a document-level
classification problem. Given a set of evaluation documents, each having
a known correct label from a closed set of labels (often referred to as
the ``gold-standard''), and a predicted label for each document from the
same set, the document-level accuracy is the proportion of documents
that are correctly labeled over the entire evaluation collection. This
is the most frequently reported metric and conveys the same information
as the error rate, which is simply the proportion of documents that are
incorrectly labeled (i.e.\ \(1 - \text{accuracy}\)).

Authors sometimes provide a per-language breakdown of results. There are two 
distinct ways in which results are generally summarized per-language: (1) precision, 
in which documents are grouped according to their predicted language; and (2) 
recall, in which documents are grouped according to what language they are 
actually written in. Earlier work has tended to only provide a breakdown based on the correct label 
(i.e.\ only reporting per-language recall). This gives us a sense of how likely
a document in any given language is to be classified correctly, but does not
give an indication of how likely a prediction for a given language is of being
correct. Under the monolingual assumption (i.e.\ each document is written in
exactly one language), this is not too much of a problem, as a false negative
for one language must also be a false positive for another language, so precision
and recall are closely linked. Nonetheless, authors have recently tended to explicitly
provide both precision and recall for clarity. It is also common practice to report
an F-score \(F\), which is the harmonic mean of precision and
recall. The F-score (also sometimes called F1-score or F-measure) was developed in IR to measure the effectiveness of retrieval with respect to a user who attaches different relative importance to
precision and recall \citep{rijsbergen1}. When used as
an evaluation metric for classification tasks, it is common to place
equal weight on precision and recall (hence ``F1''-score, in reference
to the $\beta$ hyper-parameter, which equally weights precision and
recall when $\beta=1$).

In addition to evaluating performance for each individual language,
authors have also sought to convey the relationship between
classification errors and specific sets of languages. Errors in \langid
systems are generally not random; rather, certain sets of languages are
much more likely to be confused. The typical method of conveying this
information is through the use of a confusion matrix, a tabulation of
the distribution of (predicted language, actual language) pairs.

Presenting full confusion matrices becomes problematic as the number of
languages considered increases, and as a result has become relatively
uncommon in work that covers a broader range of languages. Per-language
results are also harder to interpret as the number of languages
increases, and so it is common to present only collection-level summary
statistics. There are two conventional methods for summarizing across a
whole collection: (1) giving each document equal weight; and (2) giving
each class (i.e.\ language) equal weight. (1) is referred to as a
micro-average, and (2) as a macro-average. For \langid under the
monolingual assumption, micro-averaged precision and recall are the
same, since each instance of a false positive for one language must also
be a false negative for another language. In other words, micro-averaged
precision and recall are both simply the collection-level accuracy. On
the other hand, macro-averaged precision and recall give equal weight to
each language. In datasets where the number of documents per language is
the same, this again works out to being the collection-level
average. However, \langid research has frequently dealt with datasets
where there is a substantial skew between classes.  In such cases, the
collection-level accuracy is strongly biased towards more
heavily-represented languages. To address this issue, in work on skewed
document collections, authors tend to report both the collection-level
accuracy and the macro-averaged precision/recall/F-score, in order to
give a more complete picture of the characteristics of the method being
studied.

Whereas the notions of macro-averaged precision and recall are clearly
defined, there are two possible methods to calculate the macro-averaged
F-score. The first is to calculate it as the harmonic mean of the
macro-averaged precision and recall, and the second is to calculate it
as the arithmetic mean of the per-class F-score.

The comparability of published results is also limited by the variation
in size and source of the data used for evaluation.  In work to date,
authors have used data from a variety of different sources to evaluate
the performance of proposed solutions. Typically, data for a number of
languages is collected from a single source, and the number of languages
considered varies widely. Earlier work tended to focus on a smaller
number of Western European languages. Later work has shifted focus to
supporting larger numbers of languages simultaneously, with the work of
\cite{brown3} pushing the upper bound, reporting a language identifier
that supports over 1300 languages.  The increased size of the language
set considered is partly due to the increased availability of
language-labeled documents from novel sources such as Wikipedia and
Twitter. This supplements existing data from translations of the
Universal Declaration of Human Rights, bible translations, as well as
parallel texts from MT datasets such as OPUS and SETimes, and European
Government data such as JRC-Acquis. These factors have led to a shift
away from proprietary datasets such as the ECI multilingual corpus that
were commonly used in earlier research. As more languages are considered
simultaneously, the accuracy of \langid systems decreases. A
particularly striking illustration of this is the evaluation results by
\cite{jauhiainen6} for the logLIGA method \citep{vogel1}. \cite{vogel1}
report an accuracy of 99.8\% over tweets (averaging 80 characters) in
six European languages as opposed to the 97.9\% from the original LIGA
method. The LIGA and logLIGA implementations by \cite{jauhiainen6} have
comparable accuracy for six languages, but the accuracy for 285
languages (with 70 character test length) is only slightly over 60\% for
logLIGA and the original LIGA method is at almost 85\%. Many evaluations
are not directly comparable as the test sizes, language sets, and
hyper-parameters differ. A particularly good example is the method of
\cite{cavnar1}. The original paper reports an accuracy of 99.8\% over
eight European languages (>300 bytes test size). \cite{lui1} report an
accuracy of 68.6\% for the method over a dataset of 67 languages (500
byte test size), and \cite{jauhiainen6} report an accuracy of over 90\%
for 285 languages (25 character test size).

Separate to the question of the number and variety of languages included
are issues regarding the quantity of training data used. A number of
studies have examined the relationship between \langid accuracy and
quantity of training data through the use of learning curves. The
general finding is that \langid accuracy increases with more training
data, though there are some authors that report an optimal amount of
training data, where adding more training data decreases accuracy
thereafter \citep{ljubesic1}.  Overall, it is not clear whether there is
a universal quantity of data that is ``enough'' for any language, rather
this amount appears to be affected by the particular set of languages as
well as the domain of the data. The breakdown presented by
\cite{baldwin2} shows that with less than 100KB per language, there are
some languages where classification accuracy is near perfect, whereas
there are others where it is very poor.

Another aspect that is frequently reported on is how long a sample of
text needs to be before its language can be correctly
detected. Unsurprisingly, the general consensus is that longer samples
are easier to classify correctly. There is a strong interest in
classifying short segments of text, as certain applications naturally
involve short text documents, such as \langid of microblog messages or
search engine queries. Another area where \langid of texts as short as
one word has been investigated is in the context of dealing with
documents that contain text in more than one language, where word-level
\langid has been proposed as a possible solution (see
\secref{openissues:multilingual}). These outstanding challenges have led
to research focused specifically on \langid of shorter segments of text,
which we discuss in more detail in \secref{openissues:short}.

From a practical perspective, knowing the rate at which a \langid system
can process and classify documents is useful as it allows a practitioner
to predict the time required to process a document collection given
certain computational resources.  However, so many factors influence the
rate at which documents are processed that comparison of absolute values
across publications is largely meaningless. Instead, it is more valuable
to consider publications that compare multiple systems under controlled
conditions (same computer hardware, same evaluation data, etc.). The
most common observations are that classification times between different
algorithms can differ by orders of magnitude, and that the fastest
methods are not always the most accurate.  Beyond that, the diversity of
systems tested and the variety in the test data make it difficult to
draw further conclusions about the relative speed of algorithms.

Where explicit feature selection is used, the number of features
retained is a parameter of interest, as it affects both the memory
requirements of the \langid system and its classification rate. In
general, a smaller feature set results in a faster and more lightweight
identifier. Relatively few authors give specific details of the
relationship between the number of features selected and accuracy. A
potential reason for this is that the improvement in accuracy plateaus
with increasing feature count, though the exact number of features
required varies substantially with the method and the data used. At the
lower end of the scale, \cite{cavnar1} report that 300--400 features
per language is sufficient. Conversely \cite{jauhiainen6} found that,
for the same method, the best results for the evaluation set were
attained with 20,000 features per language.

\subsection{Corpora Used for \langid Evaluation}
\label{sec:openissues:corpora}

As discussed in \secref{standardevaluation}, the objective comparison of
different methods for \langid is difficult due to the variation in the
data that different authors have used to evaluate \langid methods.
\cite{baldwin2} emphasize this by demonstrating how the performance of
a system can vary according to the data used for evaluation.  This
implies that comparisons of results reported by different authors may
not be meaningful, as a strong result in one paper may not translate
into a strong result on the dataset used in a different paper. In other
areas of research, authors have proposed standardized corpora to allow
for the objective comparison of different methods.

Some authors have released datasets to accompany their work, to allow
for direct replication of their experiments and encourage comparison and
standardization. \tabref{datasets} lists a number of datasets that have
been released to accompany specific \langid publications. In this list,
we only include corpora that were prepared specifically for \langid
research, and that include the full text of documents. Corpora of
language-labelled Twitter messages that only provide document
identifiers are also available, but reproducing the full original corpus
is always an issue as the original Twitter messages are deleted or
otherwise made unavailable.

\begin{table}[t!]
  \centering
  \resizebox{\textwidth}{!}{%
  \begin{tabular}{lll}
  Reference & Type & Source  \\
    \hline

  \cite{baldwin2} & Multilingual (81) & Government Documents, News Texts, Wikipedia \\
  \multicolumn{3}{l}{\hspace{3mm}\url{https://github.com/varh1i/language_detection/tree/master/src/main/resources/naacl2010-langid}} \\[2mm]

  \cite{baldwin1} & Multilingual (74) & Wikipedia (synthetic multilingual docs) \\
  \multicolumn{3}{l}{\hspace{3mm}\url{http://people.eng.unimelb.edu.au/tbaldwin/etc/altw2010-langid.tgz}} \\[2mm]

  \cite{vatanen1} & Multiglingual (281) & Universal Declaration of Human Rights (UDHR) \\
  \multicolumn{3}{l}{\hspace{3mm}\url{http://research.ics.aalto.fi/cog/data/udhr/}} \\[2mm]

  \cite{tromp2} & Multilingual (6) & Twitter \\
  \multicolumn{3}{l}{\hspace{3mm}\url{http://www.win.tue.nl/~mpechen/projects/smm/LIGA_Benelearn11_dataset.zip}}\\[2mm]

  \cite{Zaidan:CallisonBurch:2011} & Arabic dialects (5) & Arabic Online Commentary (AOC)\\
  \multicolumn{3}{l}{\hspace{3mm}\url{https://github.com/sjeblee/AOC}} \\[2mm]
  
  \cite{lui1} & Multilingual & Various \\
  \multicolumn{3}{l}{\hspace{3mm}\url{http://people.eng.unimelb.edu.au/tbaldwin/etc/ijcnlp2011-langid.tgz}} \\[2mm]

  \cite{majlis2} & Multilingual (124) & Wikipedia (YALI)\\
  \multicolumn{3}{l}{\hspace{3mm}\url{http://ufal.mff.cuni.cz/tools/yali}} \\[2mm]

  \cite{tiedemann1} & Multilingual (10) & SETimes (News Texts) \\
  \multicolumn{3}{l}{\hspace{3mm}\url{http://www.nljubesic.net/resources/corpora/setimes/}} \\[2mm]

  \cite{brown2} & Multilingual (970/1279) & Bible Translations, Wikipedia \\
  \multicolumn{3}{l}{\hspace{3mm}\url{http://sourceforge.net/projects/la-strings/files/Language-Data/}} \\[2mm]

  \cite{king1} & Multilingual (30) & Web Crawl \\
  \multicolumn{3}{l}{\hspace{3mm}\url{http://www-personal.umich.edu/~benking/resources/mixed-language-annotations-release-v1.0.tgz}} \\[2mm]

  \cite{lui5} & Multilingual (65) & Twitter \\
  \multicolumn{3}{l}{\hspace{3mm}\url{http://people.eng.unimelb.edu.au/tbaldwin/data/lasm2014-twituser-v1.tgz}} \\[2mm]

  \cite{lui3} & Multilingual (156) & WikipediaMulti \\
  \multicolumn{3}{l}{\hspace{3mm}\url{http://people.eng.unimelb.edu.au/tbaldwin/etc/wikipedia-multi-v6.tgz}} \\[2mm]

  \cite{tan3} & Multilingual (22) & News Texts\\
  \multicolumn{3}{l}{\hspace{3mm}\url{http://ttg.uni-saarland.de/resources/DSLCC/}}  \\

\cite{chanda1} & Code-switching (2) & Facebook chats\\
  \multicolumn{3}{l}{\hspace{3mm}\url{https://github.com/ArunavhaChanda/Facebook-Code-Mixed-Corpus}}  \\

\cite{blodgett1} & Multilingual (70) & Twitter70\\
  \multicolumn{3}{l}{\hspace{3mm}\url{http://slanglab.cs.umass.edu/TwitterLangID/}}  \\

  \hline
  \end{tabular}
  }
  \caption{Published \langid Datasets}
  \label{tab:datasets}
\end{table}



One challenge in standardizing datasets for \langid is that the codes used to
label languages are not fully standardized, and a large proportion of labeling systems only
cover a minor portion of the languages used in the world today 
\citep{Constable:Simons:2000}. \cite{xia2} discuss this problem in 
detail, listing different language code sets, as well as the internal structure 
exhibited by some of the code sets. Some standards consider certain groups of 
``languages'' as varieties of a single macro-language, whereas others consider 
them to be discrete languages. An example of this is found in South Slavic languages,
where some language code sets refer to Serbo-Croatian, whereas others make 
distinctions between Bosnian, Serbian and Croatian 
\citep{tiedemann1}. The unclear boundaries between such languages 
make it difficult to build a reference corpus of documents for each
language, or to compare language-specific results across datasets.  

Another challenge in standardizing datasets for \langid is the great deal of variation 
that can exist between data in the same language. We examine this in greater detail in 
\secref{openissues:encoding}, where we discuss how the same language can use a number of
different orthographies, can be digitized using a number of different 
encodings, and may also exist in transliterated forms. The issue of variation 
within a language complicates the development of standardized datasets, due to
challenges in determining which variants of a language should be included. 
Since we have seen that the performance of \langid systems can vary per-domain 
\citep{baldwin2}, that \langid research is often motivated by 
target applications (see \secref{applications}), and that domain-specific 
information can be used to improve accuracy 
(see \secref{openissues:domainspecific}), it is often unsound to use a 
generic \langid dataset to develop a language identifier for a particular domain. 

A third challenge in standardizing datasets for \langid is the cost of
obtaining correctly-labeled data. Manual labeling of data is usually
prohibitively expensive, as it requires access to native speakers of all
languages that the dataset aims to include. Large quantities of raw text
data are available from sources such as web crawls or Wikipedia, but
this data is frequently mislabeled (e.g.\ most non-English Wikipedias
still include some English-language documents). In constructing corpora
from such resources, it is common to use some form of automatic \langid,
but this makes such corpora unsuitable for evaluation purposes as they
are biased towards documents that can be correctly identified by
automatic systems \citep{lui5}. Future work in this area could
investigate other means of ensuring correct gold-standard labels while
minimizing the annotation cost.

Despite these challenges, standardized datasets are critical for
replicable and comparable research in \langid.  Where a subset of data
is used from a larger collection, researchers should include details of
the specific subset, including any breakdown into training and test
data, or partitions for cross-validation. Where data from a new source
is used, justification should be given for its inclusion, as well as
some means for other researchers to replicate experiments on the same
dataset.

\subsection{\langid Shared Tasks}
\label{sec:evaluation:sharedtasks}

To address specific sub-problems in \langid, a number of shared tasks
have been organized on problems such as \langid in multilingual
documents \citep{baldwin1}, code-switched data \citep{solorio1},
discriminating between closely related languages \citep{zampieri6}, and
dialect and language variety identification in various languages
\citep{grouin10,zampieri9,rangel2,ali2}. Shared tasks are important for
\langid because they provide datasets and standardized evaluation
methods that serve as benchmarks for the \langid community. We summarize
all \langid shared tasks organized to date in \tabref{sharedtasks}.

Generally, datasets for shared tasks have been made publicly available
after the conclusion of the task, and are a good source of standardized
evaluation data. However, the shared tasks to date have tended to target
specific sub-problems in \langid, and no general, broad-coverage \langid
datasets have been compiled. Widespread interest in \langid over
closely-related languages has resulted in a number of shared tasks that
specifically tackle the issue. Some tasks have focused on varieties of a
specific language. For example, the DEFT2010 shared task \citep{grouin10}
examined varieties of French, requiring participants to classify French
documents with respect to their geographical source, in addition to the
decade in which they were published. Another example is the Arabic
Dialect Identification (``ADI'') shared task at the VarDial workshop
\citep{malmasi6,zampieri9}, and the Arabic Multi-Genre Broadcast
(``MGB'') Challenge \citep{ali2}.

Two shared tasks focused on a narrow group of languages using Twitter
data. The first was TweetLID, a shared task on \langid of Twitter
messages according to six languages in common use in Spain, namely:
Spanish, Portuguese, Catalan, English, Galician, and Basque (in order of
the number of documents in the dataset) \citep{zubiaga1,zubiaga2}. The
organizers provided almost 35,000 Twitter messages, and in addition to
the six monolingual tags, supported four additional categories:
undetermined, multilingual (i.e.\ the message contains more than one
language, without requiring the system to specify the component
languages), ambiguous (i.e.\ the message is ambiguous between two or
more of the six target languages), and other (i.e.\ the message is in a
language other than the six target languages). The second shared task
was the PAN lab on authorship profiling 2017 \citep{rangel2}. The PAN lab
on authorship profiling is held annually and historically has focused on
age, gender, and personality traits prediction in social media. In 2017
the competition introduced the inclusion of language varieties and
dialects of Arabic, English, Spanish, and Portuguese,

\begin{table}[t!]
  \centering
  \footnotesize
  \begin{tabular}{ll}
 Year -- Title & Reference  \\
    \hline

2010 -- DÉfi Fouille de Texte (DEFT) & \cite{grouin10} \\
  \hspace{3mm}\url{https://deft.limsi.fr}\\[2mm]
  
2010 -- Australasian Language Technology Workshop & \cite{baldwin1} \\
  \hspace{3mm}\url{http://www.alta.asn.au}\\[2mm]


2014 -- Twitter Language Identification Workshop at SEPLN 2014 & \cite{zubiaga1} \\
  \hspace{3mm}\url{http://komunitatea.elhuyar.org/tweetlid/?lang=en_us}\\[2mm]

2014 -- Computational Approaches to Code Switching & \cite{solorio1} \\
  \hspace{3mm}\url{http://emnlp2014.org/workshops/CodeSwitch/call.html}\\[2mm]

2014 -- First DSL Shared Task at VarDial & \cite{zampieri6} \\
  \hspace{3mm}\url{http://corporavm.uni-koeln.de/vardial/sharedtask.html}\\[2mm] 
  
2015 -- Second DSL Shared Task at VarDial  & \cite{zampieri8} \\
  \hspace{3mm}\url{http://ttg.uni-saarland.de/lt4vardial2015/dsl.html}\\[2mm] 
  
2016 -- First Arabic Dialect Identification (ADI) at VarDial & \cite{malmasi9} \\
  \hspace{3mm}\url{http://ttg.uni-saarland.de/vardial2016/dsl2016.html}\\[2mm] 

2016 -- Third DSL Shared Task at VarDial & \cite{malmasi9} \\
  \hspace{3mm}\url{http://ttg.uni-saarland.de/vardial2016/dsl2016.html}\\[2mm] 

2017 -- Second Arabic Dialect Identification (ADI) at VarDial & \cite{zampieri9} \\
  \hspace{3mm}\url{http://ttg.uni-saarland.de/vardial2017/sharedtask2017.html}\\[2mm] 

2017 -- Fourth DSL Shared Task at VarDial & \cite{zampieri9} \\
  \hspace{3mm}\url{http://ttg.uni-saarland.de/vardial2017/sharedtask2017.html}\\[2mm]   

2017 -- First German Dialect Identification (ADI) at VarDial & \cite{zampieri9} \\
  \hspace{3mm}\url{http://ttg.uni-saarland.de/vardial2017/sharedtask2017.html}\\[2mm] 

2017 -- PAN lab on Author Profiling & \cite{rangel2} \\
  \hspace{3mm}\url{http://pan.webis.de/clef17/pan17-web/author-profiling.html}\\[2mm] 
  
2017 -- Arabic Multi-Genre Broadcast (MGB) Challenge & \cite{ali2} \\
  \hspace{3mm}\url{http://www.mgb-challenge.org/arabic.html} \\
     
    \hline

  \end{tabular}
  \caption{\small{List of \langid shared tasks.}}
  \label{tab:sharedtasks}
\end{table}

More ambitiously, the four editions of the Discriminating between
Similar Languages (DSL) \citep{zampieri6,zampieri8,malmasi9,zampieri9}
shared tasks required participants to discriminate between a set of
languages in several language groups, each consisting of highly-similar
languages or national varieties of that language. The dataset, entitled
DSL Corpus Collection (``DSLCC'') \citep{tan3}, and the languages
included are summarized in \tabref{dslcc}. Historically the best-performing
systems \citep{goutte1,lui6,bestgen1} have approached the task via
hierarchical classification, first predicting the language group, then
the language within that group.

\begin{table}[!t]
\centering
\footnotesize
\begin{tabular}{lcccc}
\hline
\bf Language/Variety & \bf v1.0 (2014) & \bf v2.0/2.1 (2015) & \bf v3.0 (2016) & \bf v4.0 (2017)\\
\hline
      Bosnian 	& \checkmark & \checkmark & \checkmark & \checkmark \\
      Croatian	 & \checkmark & \checkmark & \checkmark & \checkmark \\
      Serbian	& \checkmark & \checkmark & \checkmark & \checkmark \\
\hline
	Czech	 & \checkmark & \checkmark & & \\
	Slovak		& \checkmark & \checkmark & &\\
\hline
	Indonesian	 & \checkmark & \checkmark & \checkmark & \checkmark \\
	Malay		& \checkmark & \checkmark & \checkmark & \checkmark \\
\hline
	Brazilian Portuguese	& \checkmark & \checkmark & \checkmark & \checkmark \\
	European Portuguese		& \checkmark & \checkmark & \checkmark & \checkmark \\
    Macanese Portuguese		&  & \checkmark & &\\
\hline
	Argentine Spanish	& \checkmark & \checkmark & \checkmark & \checkmark \\
	Castilian Spanish	& \checkmark & \checkmark & \checkmark & \checkmark\\
    Mexican Spanish     &  & \checkmark & \checkmark &\\
    Peruvian Spanish	& & & & \checkmark \\
\hline
	Bulgarian		&  & \checkmark &  &\\
	Macedonian	& &  \checkmark & & \\
\hline
	Canadian French		&  & & \checkmark & \checkmark \\
	Hexagonal French	& &  & \checkmark & \checkmark \\
\hline
	American English & \checkmark  & & &\\
	British English  & \checkmark & & & \\
\hline
	Persian & & & \checkmark \\
	Dari  & & & \checkmark \\
\hline
\end{tabular}
\caption{\small{DSLCC: the languages included in each version of the
    corpus collection, grouped by language similarity.}} 
\label{tab:dslcc}
\end{table}

\section{Application Areas}
\label{sec:applications}

There are various reasons to investigate \langid. Studies in \langid
approach the task from different perspectives, and with different
motivations and application goals in mind. In this section, we briefly
summarize what these motivations are, and how their specific needs
differ.

The oldest motivation for automatic \langid is perhaps in conjunction
with translation \citep{beesley1}. Automatic \langid is used
as a pre-processing step to determine what translation model to apply to an input text,
whether it be by routing to a specific human translator or by applying MT.
Such a use case is still very common, and can be seen in the Google Chrome web 
browser,\footnote{\url{http://www.google.com/chrome}} where an built-in \langid
module is used to offer MT services to the user when the detected language 
of the web page being visited differs from the user's language settings.

NLP components such as POS taggers and parsers tend to
make a strong assumption that the input text is monolingual in a given
language. Similarly to the translation case, \langid can play an obvious
role in routing documents written in different languages to NLP
components tailored to those languages. More subtle is the case of
documents with mixed multilingual content, the most commonly-occurring
instance of which is foreign inclusion, where a document is
predominantly in a single language (e.g.\ German or Japanese) but is
interspersed with words and phrases (often technical terms) from a
language such as English. For example, \cite{Alex+:2007} found that
around 6\% of word tokens in German text sourced from the Internet are
English inclusions. In the context of POS tagging, one strategy for
dealing with inclusions is to have a dedicated POS for all foreign
words, and force the POS tagger to perform both foreign inclusion
detection and POS tag these words in the target language; this is
the approach taken in the Penn POS tagset, for example
\citep{Marcus:1993}. An alternative strategy is to have an explicit
foreign inclusion detection pre-processor, and some special
handling of foreign inclusions. For example, in the context of German
parsing, \cite{Alex+:2007} used foreign inclusion predictions to
restrict the set of (German) POS tags used to form a parse tree, and
found that this approach substantially improved parser accuracy.

Another commonly-mentioned use case is for multilingual document storage
and retrieval.  A document retrieval system (such as, but not limited
to, a web search engine) may be required to index documents in multiple
languages. In such a setting, it is common to apply \langid at two
points: (1) to the documents being indexed; and (2) to the queries being
executed on the collection. Simple keyword matching techniques can be
problematic in text-based document retrieval, because the same word can
be valid in multiple languages. A classic example of such words (known
as ``false friends'') includes \textit{gift}, which in German means
``poison''. Performing \langid on both the document and the query helps
to avoid confusion between such terms, by taking advantage of the
context in which it appears in order to infer the language. This has
resulted in specific work in \langid of web pages, as well as search
engine queries. \cite{roy1} and \cite{sequeira2} give overviews of
shared tasks specifically concentrating on language labeling of
individual search query words. Having said this, in many cases, the
search query itself does a sufficiently good job of selecting documents
in a particular language, and overt \langid is often not performed in
mixed multilingual search contexts.

Automatic \langid has also been used to facilitate linguistic and other
text-based research. \cite{suzuki1} report that their motivation for
developing a language identifier was ``to find out how many web pages
are written in a particular language''. Automatic \langid has been used
in constructing web-based corpora. The Cr\'{u}bad\'{a}n project
\citep{Scannell:2007} and the Finno-Ugric Languages and the Internet
project \citep{jauhiainen4} make use of automated \langid techniques to
gather linguistic resources for under-resourced languages. Similarly,
the Online Database of INterlinear text (``ODIN'': \cite{Lewis:Xia:2010})
uses automated \langid as one of the steps in collecting interlinear
glossed text from the web for purposes of linguistic search and
bootstrapping NLP tools.

One challenge in collecting linguistic resources from the web is that
documents can be multilingual (i.e.\ contain text in more than one
language). This is problematic for standard \langid methods, which
assume that a document is written in a single language, and has prompted
research into segmenting text by language, as well as word-level
\langid, to enable extraction of linguistic resources from multilingual
documents. A number of \langid shared tasks discussed in detail in
\secref{evaluation:sharedtasks} included data from social
media. Examples are the TweetLID shared task on tweet \langid held at
SEPLN 2014 \citep{zubiaga1,zubiaga2}, the data sets used in the first and
second shared tasks on \langid in code-switched data which were
partially taken from Twitter \citep{solorio1,molina1}, and the third
edition of the DSL shared task which contained two out-of-domain test
sets consisting of tweets \citep{malmasi9}. The 5th edition of the PAN at
CLEF author profiling task included language variety identification for
tweets \citep{rangel2}. There has also been research on identifying the
language of private messages between eBay users \citep{mayer1},
presumably as a filtering step prior to more in-depth data analysis.

\section{Off-the-Shelf Language Identifiers}
\label{sec:ots}

An ``off-the-shelf'' language identifier is software that is distributed
with pre-trained models for a number of languages, so that a user is not
required to provide training data before using the system. Such a setup
is highly attractive to many end-users of automatic \langid whose main
interest is in utilizing the output of a language identifier rather than
implementing and developing the technique. To this end, a number of
off-the-shelf language identifiers have been released over time. Many
authors have evaluated these off-the-shelf identifiers, including a
recent evaluation involving 13 language identifiers which was carried
out by \cite{pawelka1}. In this section, we provide a brief summary of
open-source or otherwise free systems that are available, as well as the
key characteristics of each system. We have also included dates of when
the software has been last updated as of October 2018.

\texttt{TextCat} is the most well-known Perl implementation of the
out-of-place method, it lists models for 76 languages in its
off-the-shelf
configuration;\footnote{\url{http://odur.let.rug.nl/~vannoord/TextCat/}}
the program is not actively maintained.  \texttt{TextCat} is not the
only example of an off-the-shelf implementation of the out-of-place
method: other implementations include \texttt{libtextcat} with 76
language
models,\footnote{\url{https://software.wise-guys.nl/libtextcat/} (not
  updated since 2003)} \texttt{JTCL} with 15
languages,\footnote{\url{http://textcat.sourceforge.net} (not updated
  since first release)} and \texttt{mguesser} with 104 models for
different language-encoding
pairs.\footnote{\url{http://www.mnogosearch.org/guesser/} (not updated
  since 2008)} The main issue addressed by later implementations is
classification speed: \texttt{TextCat} is implemented in Perl and is not
optimized for speed, whereas implementations such as \texttt{libtextcat}
and \texttt{mguesser} have been specifically written to be fast and
efficient. \texttt{whatlang-rs} uses an algorithm based on character
trigrams and refers the user to the \cite{cavnar1} article. It comes
pre-trained with 83
languages.\footnote{\url{https://github.com/greyblake/whatlang-rs} (last
  updated June 2018)}

\cld is the language identifier embedded in the Google Chrome web 
browser.\footnote{\url{http://www.google.com/chrome}} It uses a NB classifier,
and script-specific classification strategies. \cld assumes that all the input is in UTF-8, and
assigns the responsibility of encoding detection and transcoding to the user.
\cld uses Unicode information to determine the script of the input.
\cld also implements a number of pre-processing heuristics to help boost
performance on its target domain (web pages), such as stripping character sequences
like \texttt{.jpg}. The standard implementation supports 83 languages, and an extended
model is also available, that supports 160 languages.\footnote{\url{https://github.com/CLD2Owners/cld2} (last updated on August 2015)}

\langdetect is a Java library that implements a language identifier
based on a NB classifier trained over character \ngrams.  The software
comes with pre-trained models for 53 languages, using data from
Wikipedia.\footnote{\url{https://github.com/shuyo/language-detection}
  (last updated on March 2014)} \langdetect makes use of a range of
normalization heuristics to improve the performance on particular
languages, including: (1) clustering of Chinese/Japanese/Korean
characters to reduce sparseness; (2) removal of ``language-independent''
characters, and other text normalization; and (3) normalization of
Arabic characters.

\ldpy is a Python implementation of the method described by
\cite{lui1}, which exploits training data for the same language across
multiple different sources of text to identify sequences of characters
that are strongly predictive of a given language, regardless of the
source of the text. This feature set is combined with a NB classifier,
and is distributed with a pre-trained model for 97 languages prepared
using data from 5 different text
sources.\footnote{\url{https://github.com/saffsd/langid.py} (last
  updated on July 2017)} \cite{lui2} provide an empirical comparison of
\ldpy to \textcat, \langdetect and \cld and find that it compares
favorably both in terms of accuracy and classification speed. There are
also implementations of the classifier component (but not the training
portion) of \ldpy in Java,
\footnote{\url{https://github.com/carrotsearch/langid-java} (last
  updated on June 2013)}
C,\footnote{\url{https://github.com/saffsd/langid.c} (last
  updated on September 2017)} and
JavaScript.\footnote{\url{https://github.com/saffsd/langid.js} (last
  updated on July 2014)}

\whatlang \citep{brown2} uses a vector-space model with per-feature weighting on
character \ngram sequences. One particular feature of \whatlang is that it uses
discriminative training in selecting features, i.e.\ it specifically makes use
of features that are strong evidence \textit{against} a particular language,
which is generally not captured by NB models. Another feature of \whatlang is that it uses
inter-string smoothing to exploit sentence-level locality in making language predictions,
under the assumption that adjacent sentences are likely to be in the same language.
\cite{brown2} reports that this substantially improves the accuracy of
the identifier. Another distinguishing feature of \whatlang is that it comes
pre-trained with data for 1400 languages, which is the highest number by a large
margin of any off-the-shelf system.\footnote{\url{https://sourceforge.net/projects/la-strings/} (last updated on February 2018)}

\texttt{whatthelang} is a recent language identifier written in Python,
which utilizes the FastText NN-based text classification algorithm. It supports 176
languages.\footnote{\url{https://github.com/indix/whatthelang} (last
  updated on November 2017)}

\yali implements an off-the-shelf classifier trained using Wikipedia
data, covering 122
languages.\footnote{\url{https://github.com/martin-majlis/YALI} (last
  updated on May 2014)} Although not described as such, the actual
classification algorithm used is a linear model, and is thus closely
related to both NB and a cosine-based vector space model.

In addition to the above-mentioned general-purpose language identifiers,
there have also been efforts to produce pre-trained language identifiers
targeted specifically at Twitter messages.  \ldig is a Twitter-specific
\langid tool with built-in models for 19
languages.\footnote{\url{https://github.com/shuyo/ldig} (last updated on
  July 2013)} It uses a document representation based on tries
\citep{Okanohara:Tsujii:2009}.  The algorithm is a LR classifier using
all possible substrings of the data, which is important to maximize the
available information from the relatively short Twitter messages.

\cite{lui5} provides a comparison of 8 off-the-shelf language
identifiers applied without re-training to Twitter messages. One issue
they report is that comparing the accuracy of off-the-shelf systems is
difficult because of the different subset of languages supported by each
system, which may also not fully cover the languages present in the
target data. The authors choose to compare accuracy over the full set of
languages, arguing that this best reflects the likely use-case of
applying an off-the-shelf \langid system to new data. They find that the
best individual systems are \cld, \ldpy and \langdetect, but that
slightly higher accuracy can be attained by a simple voting-based
ensemble classifier involving these three systems.

In addition to this, commercial or other closed-source language
identifiers and language identifier services exist, of which we name a
few. The Polyglot 3000\footnote{\url{http://www.polyglot3000.com}} and
Lextek Language
Identifier\footnote{\url{http://www.lextek.com/langid/li/}} are
standalone language identifiers for Windows. Open Xerox Language
Identifier\footnote{\url{https://open.xerox.com/Services/LanguageIdentifier}}
is a web service with available REST and SOAP APIs.

\section{Research Directions and Open Issues in \langid}
\label{sec:openissues}

Several papers have catalogued open issues in \langid
\citep{sibun1,xia2,hughes1,dasilva1,baldwin2,botha4,malmasi9}.  Some of
the issues, such as text representation (\secref{features}) and choice
of algorithm (\secref{methods}), have already been covered in detail in
this survey.  In this section, we synthesize the remaining issues into a
single section, and also add new issues that have not been discussed in
previous work. For each issue, we review related work and suggest
promising directions for future work.

\subsection{Text Preprocessing}
\label{sec:openissues:preprocessing}

Text preprocessing (also known as normalization) is an umbrella term for
techniques where an automatic transformation is applied to text before
it is presented to a classifier. The aim of such a process is to
eliminate sources of variation that are expected to be confounding
factors with respect to the target task. Text preprocessing is slightly
different from data cleaning, as data cleaning is a transformation
applied only to training data, whereas normalization is applied to both
training and test data. \cite{hughes1} raise text preprocessing as an
outstanding issue in \langid, arguing that its effects on the task have
not been sufficiently investigated. In this section, we summarize the
normalization strategies that have been proposed in the \langid
literature.

Case folding is the elimination of capitalization, replacing characters in a text with
either their lower-case or upper-case forms. Basic approaches generally map between 
\texttt{[a-z]} and \texttt{[A-Z]} in the ASCII encoding, but this approach is insufficient 
for extended Latin encodings, where diacritics must also be appropriately handled. A
resource that makes this possible is the Unicode Character
Database (UCD)\footnote{\url{http://www.unicode.org/ucd/}}
which defines uppercase, lowercase and titlecase properties for each character,
enabling automatic case folding for documents in a Unicode encoding such as UTF-8.

Range compression is the grouping of a range of characters into a single
logical set for counting purposes, and is a technique that is commonly
used to deal with the sparsity that results from character sets for
ideographic languages, such as Chinese, that may have thousands of
unique ``characters'', each of which is observed with relatively low
frequency. \cite{simoes1} use such a technique where all characters in
a given range are mapped into a single ``bucket'', and the frequency of
items in each bucket is used as a feature to represent the
document. Byte-level representations of encodings that use multi-byte
sequences to represent codepoints achieve a similar effect by
``splitting'' codepoints. In encodings such as UTF-8, the codepoints
used by a single language are usually grouped together in ``code
planes'', where each codepoint in a given code plane shares the same
upper byte.  Thus, even though the distribution over codepoints may be
quite sparse, when the byte-level representation uses byte sequences
that are shorter than the multi-byte sequence of a codepoint, the shared
upper byte will be predictive of specific languages.

Cleaning may also be applied, where heuristic rules are used to remove
some data that is perceived to hinder the accuracy of the language
identifier.  For example, \cite{suzuki1} identify HTML entities as a
candidate for removal in document cleaning, on the basis that
classifiers trained on data which does not include such entities may
drop in accuracy when applied to raw HTML documents.  \cld includes
heuristics such as expanding HTML entities, deleting digits and
punctuation, and removing SGML-like tags.  Similarly, \langdetect also
removes ``language-independent characters'' such as numbers, symbols,
URLs, and email addresses. It also removes words that are all-capitals and
tries to remove other acronyms and proper names using heuristics.

In the domain of Twitter messages, \cite{tromp2} remove
links, usernames, smilies, and hashtags (a Twitter-specific ``tagging''
feature), arguing that these entities are language independent and thus
should not feature in the model. \cite{xafopoulos1} address
\langid of web pages, and report removing HTML formatting, and applying
stopping using a small stopword list. \cite{takci5} carry out \langid
experiments on the ECI multilingual corpus and report removing
punctuation, space characters, and digits.

The idea of preprocessing text to eliminate domain-specific ``noise'' is
closely related to the idea of learning domain-independent
characteristics of a language \citep{lui1}. One difference is that
normalization is normally heuristic-driven, where a manually-specified
set of rules is used to eliminate unwanted elements of the text, whereas
domain-independent text representations are data-driven, where text from
different sources is used to identify the characteristics that a
language shares between different sources. Both approaches share
conceptual similarities with problems such as content extraction for web
pages. In essence, the aim is to isolate the components of the text that
actually represent language, and suppress the components that carry
other information. One application is the language-aware extraction of
text strings embedded in binary files, which has been shown to perform
better than conventional heuristic approaches \citep{brown1}.  Future
work in this area could focus specifically on the application of
language-aware techniques to content extraction, using models of
language to segment documents into textual and non-textual
components. Such methods could also be used to iteratively improve
\langid itself by improving the quality of training data.

\subsection{Orthography and Transliteration}
\label{sec:openissues:encoding}

\langid is further complicated when we consider that some languages
can be written in different orthographies (e.g.\ Bosnian and Serbian can be 
written in both Latin and Cyrillic script). Transliteration is another
phenomenon that has a similar effect, whereby phonetic transcriptions in another
script are produced for particular languages. These transcriptions can either
be standardized and officially sanctioned, such as the use of 
\textit{Hanyu Pinyin} for Chinese, or may also emerge irregularly and
organically as in the case of \textit{arabizi} for Arabic \citep{Yaghan:2008}.
\cite{hughes1} identify variation in the encodings and scripts used by a
given language as an open issue in \langid, pointing out that early work tended
to focus on languages written using a romanized script, and suggesting that
dealing with issues of encoding and orthography adds substantial complexity to
the task. \cite{suzuki1} discuss the relative difficulties of 
discriminating between languages that vary in any combination of encoding, 
script and language family, and give examples of pairs of languages that fall 
into each category. 

\langid across orthographies and transliteration is an area that has not received
much attention in work to date, but presents unique and interesting challenges
that are suitable targets for future research. An interesting and unexplored
question is whether it is possible to detect that documents in different
encodings or scripts are written in the same language, or what language a
text is transliterated from, without any a-priori knowledge of the
encoding or scripts used. One possible approach to this could be to take advantage
of standard orderings of alphabets in a language -- the pattern of differences 
between adjacent characters should be consistent across encodings, though
whether this is characteristic of any given language requires exploration.

\subsection{Supporting Low-Resource Languages}
\label{sec:openissues:minority}

\cite{hughes1} paint a fairly bleak picture of the support for
low-resource languages in automatic \langid. This is supported by the
arguments of \cite{xia2} who detail specific issues in building hugely
multilingual datasets. \cite{Abney:Bird:2010} also specifically called
for research into automatic \langid for low-density
languages. Ethnologue \citep{simons1} lists a total of 7099
languages. \cite{xia2} describe the Ethnologue in more detail, and
discuss the role that \langid plays in other aspects of supporting
minority languages, including detecting and cataloging resources. The
problem is circular: \langid methods are typically supervised, and need
training data for each language to be covered, but the most efficient
way to recover such data is through \langid methods.

A number of projects are ongoing with the specific aim of gathering linguistic data
from the web, targeting as broad a set of languages as possible.
One such project is the aforementioned ODIN \citep{xia3,Lewis:Xia:2010},
which aims to collect parallel snippets of text from Linguistics articles published
on the web. ODIN specifically targets articles containing Interlinear Glossed Text (IGT),
a semi-structured format for presenting text and a corresponding gloss that is 
commonly used in Linguistics.

Other projects that exist with the aim of creating text corpora for
under-resourced languages by crawling the web are the Cr\'{u}bad\'{a}n
project \citep{Scannell:2007} and SeedLing \citep{Emerson+:2014}. The
Cr\'{u}bad\'{a}n crawler uses seed data in a target language to generate
word lists that in turn are used as queries for a search engine. The
returned documents are then compared with the seed resource via an
automatic language identifier, which is used to eliminate false
positives. \cite{Scannell:2007} reports that corpora for over 400
languages have been built using this method. The SeedLing project crawls
texts from several web sources which has resulted in a total of 1451
languages from 105 language families. According to the authors, this
represents 19\% of the world's languages.

Much recent work on multilingual documents (\secref{openissues:multilingual}) 
has been done with support for minority languages as a key goal. One of the
common problems with gathering linguistic data from the web is that the data in
the target language is often embedded in a document containing data in another
language. This has spurred recent developments in text segmentation by language
and word-level \langid. \cite{lui3} present a method to detect documents that contain text in more than one language and identify
the languages present with their relative proportions in the document. The method
is evaluated on real-world data from a web crawl targeted to collect documents
for specific low-density languages.

\langid for low-resource languages is a promising area for future work.
One of the key questions that has not been clearly answered is how much data
is needed to accurately model a language for purposes of \langid.
Work to date suggests that there may not be a simple answer to this question as
accuracy varies according to the number and variety of languages modeled 
\citep{baldwin2}, as well as the diversity of data available to model a 
specific language \citep{lui1}.

\subsection{Number of Languages}
\label{sec:openissues:number}

Early research in \langid tended to focus on a very limited number of
languages (sometimes as few as 2). This situation has improved somewhat
with many current off-the-shelf language identifiers supporting on the
order of 50--100 languages (\secref{ots}). The standout in this regard
is \cite{brown3}, supporting 1311 languages in its default
configuration.  However, evaluation of the identifier of \cite{brown2}
on a different domain found that the system suffered in terms of
accuracy because it detected many languages that were not present in the
test data \citep{lui5}.

\cite{Lewis:Xia:2010} describe the construction of web crawlers
specifically targeting IGT, as well as the identification of the
languages represented in the IGT snippets. \langid for thousands of
languages from very small quantities of text is one of the issues that
they have had to tackle. They list four specific challenges for \langid
in ODIN: (1) the large number of languages; (2) ``unseen'' languages
that appear in the test data but not in training data; (3) short target
sentences; and (4) (sometimes inconsistent) transliteration into Latin
text. Their solution to this task is to take advantage of a
domain-specific feature: they assume that the name of the language that
they are extracting must appear in the document containing the IGT, and
hence treat this as a co-reference resolution problem. They report that
this approach significantly outperforms the text-based \langid approach
in this particular problem setting.

An interesting area to explore is the trade-off between the number of
languages supported and the accuracy per-language. From existing results
it is not clear if it is possible to continue increasing the number of
languages supported without adversely affecting the average accuracy,
but it would be useful to quantify if this is actually the case across a
broad range of text sources. \tabref{mostlanguages} lists the articles
where the \langid with more than 30 languages has been investigated.

\begin{table}[t!]
\centering
\footnotesize
\begin{tabular}{p{5,5cm}cp{5cm}c}
\hline
\h{Reference} & \# Lang & \h{Reference} & \# Lang\\
\hline
\cite{brown3} & 1311 &
\cite{brown2} & 1100\\
\cite{brown1} & 923 &
\cite{xia3} & c. 600\\
\cite{rodrigues1} & 372 &
\cite{king2} & 300\\
\cite{jauhiainen3} & 285 &
\cite{jauhiainen6} & 285\\
\cite{vatanen1} & 281 &
\cite{yamaguchi1} & 200+\\
\cite{cazamias1} & 200 &
\cite{chew1} & 182\\
\cite{lui7} & 143 &
\cite{kocmi1} & 136\\
\cite{majlis1} & 122 &
\cite{jauhiainen1} & 103\\
\cite{majlis2} & 90 &
\cite{lui1} & 89\\
\cite{baldwin1} & 74 &
\cite{choong1} & 68\\
\cite{baldwin2} & 67 &
\cite{lui2} & 67\\
\cite{lui5} & 65 &
\cite{goldszmidt1} & 52\\
\cite{chen1} & 48 &
\cite{lui3} & 44\\
\cite{singh1} & 39 &
\cite{Cowie+:1999} & 34\\
\cite{ludovik1} & 34 &
\cite{hammarstrom1} & 32\\
\cite{abainia1} & 32  &
\cite{king1} & 31\\
\hline
\end{tabular}
\caption{\small{Empirical evaluations with more than 30 languages.}}
\label{tab:mostlanguages}
\end{table}

\subsection{``Unseen'' Languages and Unsupervised \langid}
\label{sec:openissues:unseen}

``Unseen'' languages are languages that we do not have training data for
but may nonetheless be encountered by a \langid system when applied to
real-world data. Dealing with languages for which we do not have
training data has been identified as an issue by \cite{hughes1} and has
also been mentioned by \cite{xia3} as a specific challenge in
harvesting linguistic data from the web. \cite{elfardy1} use an
unlabeled training set with a labeled evaluation set for token-level
code switching identification between Modern Standard Arabic (MSA) and
dialectal Arabic. They utilize existing dictionaries and also a
morphological analyzer for MSA, so the system is supported by extensive
external knowledge sources. The possibility to use unannotated training
material is nonetheless a very useful feature.

Some authors have attempted to tackle the unseen language problem
through attempts at unsupervised labeling of text by
language. \cite{mather1} uses an unsupervised clustering algorithm to
separate a multilingual corpus into groups corresponding to
languages. She uses singular value decomposition (SVD) to first identify
the words that discriminate between documents and then to separate the
terms into highly correlating groups. The documents grouped together by
these discriminating terms are merged and the process is repeated until
the wanted number of groups (corresponding to languages) is
reached. \cite{biemann1} also presents an approach to unseen language
problem, building graphs of co-occurrences of words in sentences, and
then partitioning the graph using a custom graph-clustering algorithm
which labels each word in the cluster with a single label. The number of
labels is initialized to be the same as the number of words, and
decreases as the algorithm is recursively applied. After a small number
of iterations (the authors report 20), the labels become relatively
stable and can be interpreted as cluster labels. Smaller clusters are
then discarded, and the remaining clusters are interpreted as groups of
words for each language. \cite{shiells1} compared the Chinese Whispers
algorithm of \cite{biemann1} and Graclus clustering on unsupervised
Tweet \langid. They conclude that Chinese Whispers is better suited
to \langid. \cite{selamat2} used Fuzzy ART NNs for unsupervised language
clustering for documents in Arabic, Persian, and Urdu. In Fuzzy ART, the
clusters are also dynamically updated during the identification process.

\cite{amine1} also tackle the unseen language problem through
clustering.  They use a character \ngram representation for text, and a
clustering algorithm that consists of an initial $k$-means phase,
followed by particle-swarm optimization. This produces a large number of
small clusters, which are then labeled by language through a separate
step. \cite{wan1} used co-occurrences of words with $k$-means
clustering in word-level unsupervised \langid. They used a Dirichlet
process Gaussian mixture model (``DPGMM''), a non-parametric variant of a
GMM, to automatically determine the number of clusters, and manually
labeled the language of each cluster. \cite{poulston1} also
used $k$-means clustering, and \cite{alfter1} used the $x$-means
clustering algorithm in a custom framework. \cite{lin2} utilized
unlabeled data to improve their \langid system by using a CRF
autoencoder, unsupervised word embeddings, and word lists.

A different partial solution to the issue of unseen languages is to
design the classifier to be able to output ``unknown'' as a prediction
for language. This helps to alleviate one of the problems commonly
associated with the presence of unseen languages -- classifiers without
an ``unknown'' facility are forced to pick a language for each document,
and in the case of unseen languages, the choice may be arbitrary and
unpredictable \citep{biemann1}. When \langid is used for filtering
purposes, i.e.\ to select documents in a single language, this
mislabeling can introduce substantial noise into the data extracted;
furthermore, it does not matter what or how many unseen languages there
are, as long as they are consistently rejected. Therefore the
``unknown'' output provides an adequate solution to the unseen language
problem for purposes of filtering.

The easiest way to implement unknown language detection is through
thresholding. Most systems internally compute a score for each language
for an unknown text, so thresholding can be applied either with a global
threshold \citep{Cowie+:1999}, a per-language threshold \citep{suzuki1},
or by comparing the score for the top-scoring $N$-languages. The problem
of unseen languages and open-set recognition was also considered by
\cite{malmasi2}, \cite{zampieri7}, and
\cite{malmasi6}. \cite{malmasi6} experiments with one-class
classification (``OCC'') and reaches an F-score on 98.9 using OC-SVMs (SVMs
trained only with data from one language) to discriminate between 10
languages.

Another possible method for unknown language detection that has not been
explored extensively in the literature, is the use of non-parametric
mixture models based on Hierarchical Dirichlet Processes (``HDP''). Such
models have been successful in topic modeling, where an outstanding
issue with the popular LDA model is the need to specify the number of
topics in advance. \cite{lui3} introduced an approach to detecting
multilingual documents that uses a model very similar to LDA, where
languages are analogous to topics in the LDA model. Using a similar
analogy, an HDP-based model may be able to detect documents that are
written in a language that is not currently modeled by the
system. \cite{voss2} used LDA to cluster unannotated tweets. Recently
\cite{zhang1} used LDA in unsupervised sentence-level \langid. They
manually identified the languages of the topics created with LDA. If
there were more topics than languages then the topics in the same
language were merged.

Filtering, a task that we mentioned earlier in this section, is a very
common application of \langid, and it is therefore surprising that there
is little research on filtering for specific languages. Filtering is a
limit case of \langid with unseen languages, where all languages but one
can be considered unknown. Future work could examine how useful
different types of negative evidence are for filtering -- if we want to
detect English documents, e.g., are there empirical advantages in having
distinct models of Italian and German (even if we don't care about the
distinction between the two languages), or can we group them all
together in a single ``negative'' class? Are we better off including as
many languages as possible in the negative class, or can we safely
exclude some?

\subsection{Multilingual Documents}
\label{sec:openissues:multilingual}

Multilingual documents are documents that contain text in more than one
language. In constructing the hrWac corpus, \cite{stupar1} found that
4\% of the documents they collected contained text in more than one
language. \cite{martins1} report that web pages in many languages
contain formulaic strings in English that do not actually contribute to
the content of the page, but may nonetheless confound attempts to
identify multilingual documents. Recent research has investigated how to
make use of multilingual documents from sources such as web crawls
\citep{king1}, forum posts \citep{nguyen1}, and microblog messages
\citep{Ling+:2013}. However, most \langid methods assume that a document
contains text from a single language, and so are not directly applicable
to multilingual documents.

Handling of multilingual documents has been named as an open research
question \citep{hughes1}. Most NLP techniques presuppose monolingual
input data, so inclusion of data in foreign languages introduces noise,
and can degrade the performance of NLP systems. Automatic detection of
multilingual documents can be used as a pre-filtering step to improve
the quality of input data. Detecting multilingual documents is also
important for acquiring linguistic data from the web, and has
applications in mining bilingual texts for statistical MT from online
resources \citep{Ling+:2013}, or to study code-switching phenomena in
online communications.  There has also been interest in extracting text
resources for low-density languages from multilingual web pages
containing both the low-density language and another language such as
English.

The need to handle multilingual documents has prompted researchers to
revisit the granularity of \langid. Many researchers consider
document-level \langid to be relatively easy, and that sentence-level
and word-level \langid are more suitable targets for further
research. However, word-level and sentence-level tokenization are not
language-independent tasks, and for some languages are substantially
harder than others \citep{Peng+:2004}.

\linguini \citep{prager1} is a language identifier that supports 
identification of multilingual documents. The system is based on a vector space 
model using cosine similarity. \langid for multilingual 
documents is performed through the use of \textit{virtual mixed languages}. 
\cite{prager1} shows how to construct vectors representative of particular 
combinations of languages independent of the relative proportions, and proposes 
a method for choosing combinations of languages to consider for any given 
document.  One weakness of this approach is that for exhaustive coverage, this 
method is factorial in the number of languages, and as such intractable for a 
large set of languages. Furthermore, calculating the parameters for the virtual 
mixed languages becomes infeasibly complex for mixtures of more than 3 
languages.

As mentioned previously, \cite{lui3} propose an LDA-inspired \langid
method for multilingual documents that is able to identify that a
document is multilingual, identify the languages present and estimate
the relative proportions of the document written in each language.  To
remove the need to specify the number of topics (or in this case,
languages) in advance, \cite{lui3} use a greedy heuristic that attempts
to find the subset of languages that maximizes the posterior probability
of a target document. One advantage of this approach is that it is not
constrained to 3-language combinations like the method of
\cite{prager1}. Language set identification has also been considered by
\cite{suzuki1}, \cite{jauhiainen3}, and \cite{pla1,pla2}.

To encourage further research on \langid for multilingual documents, in
the aforementioned shared task hosted by the Australiasian Language
Technology Workshop 2010, discussed in \secref{evaluation:sharedtasks},
participants were required to predict the language(s) present in a
held-out test set containing monolingual and bilingual documents
\citep{baldwin1}. The dataset was prepared using data from Wikipedia, and
bilingual documents were produced using a segment from an article in one
language and a segment from the equivalent article in another
language. Equivalence between articles was determined using the
cross-language links embedded within each Wikipedia
article.\footnote{Note that such articles are not necessarily direct
  translations but rather articles about the same topic written in
  different languages.} The winning entry \citep{Tran+:2010} first built
monolingual models from multilingual training data, and then applied
them to a chunked version of the test data, making the final prediction
a function of the prediction over chunks.

Another approach to handling multilingual documents is to attempt to
segment them into contiguous monolingual segments. In addition to
identifying the languages present, this requires identifying the
locations of boundaries in the text which mark the transition from one
language to another.  Several methods for supervised language
segmentation have been proposed.  \cite{Cowie+:1999} generalized a
\langid algorithm for monolingual documents by adding a dynamic
programming algorithm based on a simple Markov model of multilingual
documents. More recently, multilingual \langid algorithms have also been
presented by \cite{jhamtani1}, \cite{minocha1}, \cite{petho1},
\cite{ullman1}, and \cite{king4}.

\subsection{Short Texts}
\label{sec:openissues:short}

\langid of short strings is known to be challenging for existing \langid
techniques. \cite{mandl1} tested four different classification methods,
and found that all have substantially lower accuracy when applied to
texts of 25 characters compared with texts of 125 characters. These
findings were later strengthened, for example, by \cite{vatanen1} and
\cite{jauhiainen6}.

\cite{hammarstrom1} describes a method specifically targeted at short
texts that augments a dictionary with an affix table, which was tested
over synthetic data derived from a parallel bible corpus.
\cite{vatanen1} focus on messages of 5--21 characters, using \ngram
language models over data drawn the from Universal Declaration of Human
Rights (UDHR). We would expect that generic methods for \langid of short
texts should be effective in any domain where short texts are found,
such as search engine queries or microblog messages. However,
\cite{hammarstrom1} and \cite{vatanen1} both only test their systems
in a single domain: bible texts in the former case, and texts from the
UDHR in the latter case. Other research has shown that \langid results
do not trivially generalize across domains \citep{baldwin2}, and found
that \langid in UDHR documents is relatively easy \citep{yamaguchi1}. For
both bible and UDHR data, we expect that the linguistic content is
relatively grammatical and well-formed, an expectation that does not
carry across to domains such as search engine queries and
microblogs. Another ``short text'' domain where \langid has been studied
is \langid of proper names. \cite{hakkinen1} identify this as an
issue. \cite{Konstantopoulos:2007} found that \langid of names is more
accurate than \langid of generic words of equivalent length.

\cite{bergsma1} raise an important criticism of \langid work on Twitter
messages to date: only a small number of European languages has been
considered. \cite{bergsma1} expand the scope of \langid for Twitter,
covering nine languages across Cyrillic, Arabic and Devanagari scripts.
\cite{lui5} expand the evaluation further, introducing a dataset of
language-labeled Twitter messages across 65 languages constructed using
a semi-automatic method that leverages user identity to avoid inducing
a bias in the evaluation set towards messages that existing systems are
able to identify correctly. \cite{lui5} also test a 1300-language model
based on \cite{brown2}, but find that it performs relatively poorly in
the target domain due to a tendency to over-predict low-resource
languages.

Work has also been done on \langid of single words in a document, where
the task is to label each word in the document with a specific language.
Work to date in this area has assumed that word tokenization can be
carried out on the basis of whitespace. \cite{singh2} explore
word-level \langid in the context of segmenting a multilingual document
into monolingual segments. Other work has assumed that the languages
present in the document are known in advance.

Conditional random fields (``CRFs'': \cite{lafferty1}) are a sequence
labeling method most often used in \langid for labeling the language of
individual words in a multilingual text. CRFs can be thought of as a
finite state model with probabilistic transition probabilities optimised
over pre-defined cliques. They can use any observations made from the
test document as features, including language labels given by
monolingual language identifiers for words. \cite{king1} used a CRF
trained with generalized expectation criteria, and found it to be the
most accurate of all methods tested (NB, LR, HMM, CRF) at word-level
\langid. \cite{king1} introduce a technique to estimate the parameters
using only monolingual data, an important consideration as there is no
readily-available collection of manually-labeled multilingual documents
with word-level annotations. \cite{nguyen1} present a two-pass approach
to processing Turkish-Dutch bilingual documents, where the first pass
labels each word independently and the second pass uses the local
context of a word to further refine the predictions. \cite{nguyen1}
achieved 97,6\% accuracy on distinguishing between the two languages
using a linear-chain CRF. \cite{clematide1} are the only ones so far to
use a CRF for \langid of monolingual texts. With a CRF, they attained a
higher F-score in German dialect identification than NB or an ensemble
consisting of NB, CRF, and SVM. Lately CRFs were also used for \langid
by \cite{dongen1} and \cite{samih3}. \cite{giwa1} investigate \langid
of individual words in the context of code switching. They find that
smoothing of \ngram models substantially improves accuracy of a language
identifier based on a NB classifier when applied to individual words.

\subsection{Similar Languages, Language Varieties, and Dialects}
\label{sec:openissues:related}

While one line of research into \langid has focused on pushing the
boundaries of how many languages are supported simultaneously by a
single system \citep{xia2,brown1,brown2}, another has taken a
complementary path and focused on \langid in groups of similar
languages. Research in this area typically does not make a distinction
between languages, varieties and dialects, because such terminological
differences tend to be politically rather than linguistically motivated
\citep{clyne92,xia2,zampieri4}, and from an NLP perspective the
challenges faced are very similar.

\langid for closely-related languages, language varieties, and dialects
has been studied for Malay--Indonesian \citep{bali1}, Indian languages
\citep{murthy1}, South Slavic languages
\citep{ljubesic1,tiedemann1,ljubesic4,ljubesic2}, Serbo-Croatian dialects
\citep{andelka2013}, English varieties \citep{lui4,simaki1},
Dutch--Flemish \citep{lee3}, Dutch dialects (including a temporal
dimension) \citep{trieschnigg1}, German Dialects \citep{hollenstein1}
Mainland--Singaporean--Taiwanese Chinese \citep{Huang:Lee:2008},
Portuguese varieties \citep{zampieri4,zampieri10}, Spanish varieties
\citep{zampieri3,maier1}, French varieties
\citep{mokhov10,mokhov11,Diwersy+:2014}, languages of the Iberian
Peninsula \citep{zubiaga1}, Romanian dialects \citep{ciobanu2}, and Arabic
dialects \citep{elfardy2,zaidan1,tillmann1,sadat2,wray1}, the last of
which we discuss in more detail in this section. As to off-the-shelf
tools which can identify closely-related languages, \cite{zampieri5}
released a \langid system trained to identify 27 languages, including 10
language varieties. Closely-related languages, language varieties, and
dialects have also been the focus of a number of shared tasks in recent
years as discussed in \secref{evaluation:sharedtasks}.

Similar languages are a known problem for existing language identifiers
\citep{bali1,zampieri1}.  \cite{suzuki1} identify language pairs from
the same language family that also share a common script and the same
encoding, as the most difficult to discriminate.  \cite{tiedemann1}
report that \textcat achieves only 45\% accuracy when trained and tested
on 3-way Bosnian/Serbian/Croatian dataset. \cite{lui4} found that
\langid methods are not competitive with conventional word-based
document categorization methods in distinguishing between national
varieties of English.  \cite{bali1} reports that a character trigram
model is able to distinguish Malay/Indonesian from English, French,
German, and Dutch, but handcrafted rules are needed to distinguish
between Malay and Indonesian. One kind of rule is the use of ``exclusive
words'' that are known to occur in only one of the languages. A similar
idea is used by \cite{tiedemann1}, in automatically learning a
``blacklist'' of words that have a strong negative correlation with a
language -- i.e.\ their presence implies that the text is \emph{not}
written in a particular language. In doing so, they achieve an overall
accuracy of 98\%, far surpassing the 45\% of \textcat.  \cite{brown2}
also adopts such ``discriminative training'' to make use of negative
evidence in \langid.

\cite{zampieri1} observed that general-purpose approaches to \langid
typically use a character \ngram representation of text, but successful
approaches for closely-related languages, varieties, and dialects seem
to favor a word-based representation or higher-order \ngrams (e.g.\
4-grams, 5-grams, and even 6-grams) that often cover whole words
\citep{Huang:Lee:2008,tiedemann1,lui4,goutte3}. The study compared
character \ngrams with word-based representations for \langid over
varieties of Spanish, Portuguese and French, and found that word-level
models performed better for varieties of Spanish, but character \ngram
models perform better in the case of Portuguese and French.

To train accurate and robust \langid systems that discriminate between
language varieties or similar languages, models should ideally be able
to capture not only lexical but more abstract systemic differences
between languages. One way to achieve this, is by using features that
use de-lexicalized text representations (e.g.\ by substituting named
entities or content words by placeholders), or at a higher level of
abstraction, using POS tags or other morphosyntactic information
\citep{zampieri3,lui6,bestgen1}, or even adversarial machine learning to
modify the learned representations to remove such artefacts
\citep{Li+:2018a}. Finally, an interesting research direction could be to
combine work on closely-related languages with the analysis of regional
or dialectal differences in language use
\citep{Peirsman+:2010,anstein13,Doyle:2014,Diwersy+:2014}.

In recent years, there has been a significant increase of interest in
the computational processing of Arabic. This is evidenced by a number of
research papers in several NLP tasks and applications including the
identification/discrimination of Arabic dialects
\citep{elfardy2,zaidan1}. Arabic is particularly interesting for
researchers interested in language variation due to the fact that the
language is often in a diaglossic situation, in which the standard form
(Modern Standard Arabic or ``MSA'') coexists with several regional
dialects which are used in everyday communication.

Among the studies published on the topic of Arabic \langid,
\cite{elfardy2} proposed a supervised approach to distinguish between
MSA and Egyptian Arabic at the sentence level,
and achieved up to 85.5\% accuracy over an Arabic online commentary
dataset \citep{Zaidan:CallisonBurch:2011}. \cite{tillmann1} achieved
higher results over the same dataset using a linear-kernel SVM
classifier.

\cite{zaidan1} compiled a dataset containing MSA,
Egyptian Arabic, Gulf Arabic and Levantine Arabic, and used it to
investigate three classification tasks: (1) MSA and dialectal Arabic;
(2) four-way classification -- MSA, Egyptian Arabic, Gulf Arabic, and
Levantine Arabic; and (3) three-way classification -- Egyptian Arabic,
Gulf Arabic, and Levantine Arabic.

\cite{salloum1} explores the use of sentence-level Arabic
dialect identification as a pre-processor for MT, in customizing the
selection of the MT model used to translate a given sentence to the
dialect it uses. In performing dialect-specific MT, the authors achieve
an improvement of 1.0\% BLEU score compared with a baseline system
which does not differentiate between Arabic dialects. 

Finally, in addition to the above-mentioned dataset of
\cite{Zaidan:CallisonBurch:2011}, there are a number of notable
multi-dialect corpora of Arabic: a multi-dialect corpus of broadcast
speeches used in the ADI shared task \citep{ali1}; a multi-dialect corpus
of (informal) written Arabic containing newspaper comments and Twitter
data \citep{cotterell1}; a parallel corpus of 2,000 sentences in MSA,
Egyptian Arabic, Tunisian Arabic, Jordanian Arabic, Palestinian Arabic,
and Syrian Arabic, in addition to English \citep{bouamor2014}; a corpus
of sentences in 18 Arabic dialects (corresponding to 18 different
Arabic-speaking countries) based on data manually sourced from web
forums \citep{sadat2}; and finally two recently compiled multi-dialect
corpora containing microblog posts from Twitter
\citep{elgabou1,alshutayri2}.

While not specifically targeted at identifying language varieties,
\cite{jurgens1} made the critical observation that when naively
trained, \langid systems tend to perform most poorly over language
varieties from the lowest socio-economic demographics (focusing
particularly on the case of English), as they tend to be most
under-represented in training corpora. If, as a research community, we
are interested in the social equitability of our systems, it is critical
that we develop datasets that are truly representative of the global
population, to better quantify and remove this effect. To this end,
\cite{jurgens1} detail a method for constructing a more representative
dataset, and demonstrate the impact of training on such a dataset in
terms of alleviating socio-economic bias.

\subsection{Domain-specific \langid}
\label{sec:openissues:domainspecific}

One approach to \langid is to build a generic language identifier that
aims to correctly identify the language of a text without any
information about the source of the text. Some work has specifically
targeted \langid across multiple domains, learning characteristics of
languages that are consistent between different sources of text
\citep{lui1}. However, there are often domain-specific features that are
useful for identifying the language of a text.  In this survey, our
primary focus has been on \langid of digitally-encoded text, using only
the text itself as evidence on which to base the prediction of the
language. Within a text, there can sometimes be domain-specific
peculiarities that can be used for \langid. For example, \cite{mayer1}
investigates \langid of user-to-user messages in the eBay e-commerce
portal.  He finds that using only the first two and last two words of a
message is sufficient for identifying the language of a message.

\section{Conclusions}
\label{sec:conclusions}

This article has presented a comprehensive survey on language
identification of digitally-encoded text. We have shown that \langid is
a rich, complex, and multi-faceted problem that has engaged a wide
variety of research communities. \langid accuracy is critical as it is
often the first step in longer text processing pipelines, so errors made
in \langid will propagate and degrade the performance of later
stages. Under controlled conditions, such as limiting the number of
languages to a small set of Western European languages and using long,
grammatical, and structured text such as government documents as
training data, it is possible to achieve near-perfect accuracy. This led
many researchers to consider \langid a solved problem, as argued by
\cite{mcnamee1}. However, \langid becomes much harder when taking into
account the peculiarities of real-world data, such as very short
documents (e.g.\ search engine queries), non-linguistic ``noise'' (e.g.\
HTML markup), non-standard use of language (e.g.\ as seen in social
media data), and mixed-language documents (e.g.\ forum posts in
multilingual web forums).

Modern approaches to \langid are generally data-driven and are based on
comparing new documents with models of each target language learned from
data.  The types of models as well as the sources of training data used
in the literature are diverse, and work to date has not compared and
evaluated these in a systematic manner, making it difficult to draw
broader conclusions about what the ``best'' method for \langid actually
is. We have attempted to synthesize results to date to identify a set of
\langid ``best practices'', but these should be treated as guidelines
and should always be considered in the broader context of a target
application.

Existing work on \langid serves to illustrate that the scope and depth
of the problem are much greater than they may first appear. In
\secref{openissues}, we discussed open issues in \langid, identifying
the key challenges, and outlining opportunities for future research. Far
from being a solved problem, aspects of \langid make it an archetypal
learning task with subtleties that could be tackled by future work on
supervised learning, representation learning, multi-task learning,
domain adaptation, multi-label classification and other subfields of
machine learning. We hope that this paper can serve as a reference point
for future work in the area, both for providing insight into work to
date, as well as pointing towards the key aspects that merit further
investigation.

\appendix

\acks{This research was supported in part by the Australian Research
  Council, the Kone Foundation and the Academy of Finland. We would like
  to thank Kimmo Koskenniemi for many valuable discussions and comments
  concerning the early phases of the features and the methods sections.}

\vskip 0.2in

\bibliographystyle{plainnat}
\bibliography{strings,LanguageIdentification}


\end{document}

%% file: macros.tex
\newcommand{\secref}[2][]{Section#1~\ref{sec:#2}}
\newcommand{\ssecref}[1]{\S\ref{sec:#1}}

\newcommand{\tabref}[2][]{Table#1~\ref{tab:#2}}
\newcommand{\eqnref}[2][]{Equation#1~\ref{eqn:#2}}

\newcommand{\langid}{LI\xspace}

\newcommand{\ngram}{\ensuremath{n}-gram\xspace}
\newcommand{\ngrams}{\ensuremath{n}-grams\xspace}

\newcommand{\lex}[1]{\textit{#1}\xspace}

\newcommand{\h}[1]{\textbf{#1}\xspace}

\newcommand{\bestone}{***\xspace}
\newcommand{\besttwo}{**\xspace}
\newcommand{\bestOther}{*\xspace}
\newcommand{\bestother}{*\xspace}

\DeclareMathOperator\rank{rank}

\newcommand{\classifier}[1]{\texttt{#1}\xspace}
\newcommand{\textcat}{\classifier{TextCat}}
\newcommand{\linguini}{\classifier{Linguini}}
\newcommand{\langdetect}{\classifier{LangDetect}}
\newcommand{\cld}{\classifier{ChromeCLD}}
\newcommand{\ldig}{\classifier{LDIG}}
\newcommand{\ldpy}{\classifier{langid.py}}

\newcommand{\whatlang}{\classifier{whatlang}}
\newcommand{\yali}{\classifier{YALI}}